\documentclass[runningheads]{llncs}
\usepackage[T1]{fontenc}

\usepackage{graphicx}

\usepackage{href-ul}
\usepackage{color}

\usepackage{amsmath} 
\usepackage{amssymb} 
\usepackage{amsfonts}
\usepackage{bm}      

\usepackage[hang,small,bf]{caption} 
\usepackage[subrefformat=parens]{subcaption}

\usepackage{tikz}

\usepackage{booktabs}

\usepackage{xspace}

\usepackage{musicography}
\def\doublebemol{\musDoubleFlat}
\def\bemol{\musFlat}
\def\becarre{\musNatural}
\def\diese{\musSharp}
\def\doublediese{\musDoubleSharp}

\def\<#1>{\langle #1 \rangle}

\def\Ap{\mathsf{A}}
\def\Bp{\mathsf{B}}
\def\Cp{\mathsf{C}}
\def\Dp{\mathsf{D}}
\def\Ep{\mathsf{E}}
\def\Fp{\mathsf{F}}
\def\Gp{\mathsf{G}}

\def\Ass{\mathsf{A}\doublediese}
\def\Bss{\mathsf{B}\doublediese}
\def\Css{\mathsf{C}\doublediese}
\def\Dss{\mathsf{D}\doublediese}
\def\Ess{\mathsf{E}\doublediese}
\def\Fss{\mathsf{F}\doublediese}
\def\Gss{\mathsf{G}\doublediese}

\def\As{\mathsf{A}\diese} 
\def\Bs{\mathsf{B}\diese}
\def\Cs{\mathsf{C}\diese}
\def\Ds{\mathsf{D}\diese}
\def\Es{\mathsf{E}\diese}
\def\Fs{\mathsf{F}\diese}
\def\Gs{\mathsf{G}\diese}

\def\An{\mathsf{A}\becarre} 
\def\Bn{\mathsf{B}\becarre}
\def\Cn{\mathsf{C}\becarre}
\def\Dn{\mathsf{D}\becarre}

\def\Fn{\mathsf{F}\becarre}
\def\Gn{\mathsf{G}\becarre}

\def\Af{\mathsf{A}\bemol} 
\def\Bf{\mathsf{B}_{\bemol}}
\def\Cf{\mathsf{C}\bemol}
\def\Df{\mathsf{D}\bemol}
\def\Ef{\mathsf{E}\bemol}
\def\Ff{\mathsf{F}\bemol}
\def\Gf{\mathsf{G}\bemol}

\def\Aff{\mathsf{A}\doublebemol}
\def\Bff{\mathsf{B}\doublebemol}
\def\Cff{\mathsf{C}\doublebemol}
\def\Dff{\mathsf{D}\doublebemol}
\def\Eff{\mathsf{E}\doublebemol}
\def\Fff{\mathsf{F}\doublebemol}
\def\Gff{\mathsf{G}\doublebemol}

\newcommand{\pc}{\mathit{pc}}

\DeclareMathOperator*{\argmin}{arg\,min}

\def\ie{\textit{i.e.},\xspace}
\def\eg{\textit{e.g.},\xspace}
\def\wrt{\textit{wrt}\xspace}
\def\etc{\textit{etc}\xspace}

\begin{document}
\title{Pitch Spelling Jazz Lead Sheets,\\ Solo Transcriptions, Classical Piano and Monophonic Scores}
\titlerunning{Pitch Spelling jazz lead sheets, soli and classical piano}
%
\author{Augustin Bouquillard\inst{1}\orcidID{0009-0003-0371-3196}\\ \and
Florent Jacquemard\inst{2,3}\orcidID{0000-0003-2269-7550}}
\authorrunning{A. Bouquillard and F. Jacquemard}
%
\institute{École polytechnique, Palaiseau, France\\
\email{augustin.bouquillard@polytechnique.org}\\
\and
INRIA, Paris, France\\
\email{florent.jacquemard@inria.fr} \url{https://florent-jacquemard.github.io}}
\maketitle              
\begin{abstract}
We present an algorithm for pitch spelling and key estimation.
Given an input in MIDI-like format, containing information on note pitches 
(expressed in semitones relative to the lowest reference note) and bar boundaries, 
it estimates the appropriate note names, 
a global Key Signature, 
and a local scale for each bar.
This related information elements are evaluated jointly during two stages of optimisation. 
During an initial ‘modal’ stage, a probable scale is proposed for each bar, 
minimising the number of accidentals to be printed in the printed score with a shortest-path search. 
Then, during a second stage called ‘tonal’, these local scales are used to estimate the Key Signature 
and note names that would result in the best musical notation for the entire piece.

We present evaluations conducted on datasets comprising a variety of digital musical scores:
jazz lead sheets taken from the Real Book, transcriptions of recordings of jazz soli and bass lines,
traditional tunes,
as well as classical scores for piano and monophonic instruments.

Our procedure was originally designed for use in music transcription, 
specifically for building digital collections of jazz solos transcribed from audio recordings, 
for the purposes of music analysis, teaching and the preservation of cultural heritage.
This method should also prove useful for other tasks related to the processing of musical notation.
Furthermore, to this end, we have defined new distances between various common jazz scales, 
which may be of some interest to musicological studies.

\end{abstract}


\section{Introduction}\label{sec:intro}
Pitches can be expressed in various ways in musical notation. 
In their simplest form, according to the MIDI standard, 
they are encoded as an integral number of semitones, 
corresponding to a key number on a digital keyboard. 
Their representation is significantly more complex in traditional Western musical notation (CWN), 
where each pitch can be designated in several ways, 
using a note name, in $\Cp$, $\Dp$, $\Ep$, $\Fp$, $\Gp$, $\Ap$, $\Bp$, 
and an accidental mark in 
\musDoubleFlat, 
\musFlat, 
\musNatural, 
\musSharp, 
\musDoubleSharp, 
acting as a pitch class modifier.
The choice of a note name (\eg $\Bf$ vs $\As$ vs $\Cff$) depends on the musical context in which it appears: 
the key (Key Signature, KS) of the piece, 
the structure of the voice leading (ascending or descending melodic movements), the harmonic context, etc....

The problem of Pitch Spelling (PS) involves choosing appropriate names 
to denote pitch values that are initially expressed as absolute numbers of semitones.
Regarded as a sub-task of musical transcription (the conversion of a performance into a score in CWN), 
this problem is crucial for a number of reasons.\\
\noindent 
Firstly, for the sake of readability: 
choosing note names can reduce the number of accidentals printed on the score. 
At the global level, a Key Signature (KS) is defined for the entire piece (or part of it), 
specifying between 0 and 7 sharps or flats which are printed only at the start of each staff 
and apply by default to all notes in the staff.\\
\noindent 
Secondly, at the local level, the presence of printed accidentals, which do not appear in the global Key Signature, 
indicates the tonal function of the notes and signals local modulations. 
Moreover, the same chord can be written differently depending on whether its harmonic nature or its tonal function is to be emphasised, 
sometimes independently of neighbouring melodic movements (notable examples include the first of Chopin’s Ballades or the Tristan chord). 
In this sense, the choice of note names 
reveals the composer’s creative intent,  
beyond the practical benefit of making it easier for a musician to read a piece. 

In summary, in the tonal system, establishing an \emph{global key} 
(or KS) for a piece designates default, 
preferred note names and accidentals, 
thereby defining the piece’s tonal context from the outset. 
The presence of other accidentals (outside the Key Signature) indicates the tonal function of the notes, 
which helps to better understand the composer’s intention, particularly with regard to changes in \emph{local key}.  
There is therefore a strong interdependence between, 
on the one hand, the problem of Pitch Spelling, 
and, on the other hand, the problems of estimating local and global key or KS (KE).

Several PS algorithms have been proposed in the literature.
Many of them have been designed according to musicological criteria, 
such as the analysis of voice-leading, 
interval relationships and local keys~\cite{Chew05spiral,Meredith06ps13,Temperley04cognition},
or a principle of parsimony 
(minimisation of the number of accidentals)~\cite{Bouquillard:hal-04458185,Cambouropoulos03pitch}.
Some involve relevant intermediate data structures, 
such as the Euler lattice~\cite{Honingh09compactness} 
or weighted oriented graphs~\cite{Wetherfield20minimum}, 
in order to reduce PS to optimisation problems. 
Other approaches involve training statistical models such as HMMs~\cite{Teodoru07pitch}
or RNNs~\cite{Foscarin21PKspell}
on datasets consisting of digital musical scores. 
%
Most of above cited PS systems have been evaluated 
on corpora from classical or (less often) romantic repertoire, 
demonstrating their ability to guess reference spellings
with a high degree of accuracy.
The last cited system, PKspell~\cite{Foscarin21PKspell}, 
has obtained state-of-the-art results on an iconic benchmark called Musedata, 
proposed by D. Meredith~\cite{Meredith06ps13}, 
made of 216 works by Baroque and Classical composers.
To our knowledge, the applicability of the different approaches to jazz music,  
for example by trying to take into account the use of jazz modes,
has not been studied so far.

\smallskip 
In this work, we propose an algorithm  
addressing the PS problem, as well as, jointly, 
the estimation of a global KS and a local tonalities. 
For the latter, we consider a list of candidate modes 
including~9 diatonic modes (major, natural minor, harmonic minor, as well as the 6 other church modes), 
and two blues modes.
That makes our procedure suitable for processing jazz music, 
in particular transcriptions of improvised soli.

An originality of our approach is the use of a 1-bar\footnote{We use the term bar  to refer to a measure throughout this article.} 
window for estimating note names and local keys. 
The choice of a 1-bar duration is not arbitrary: according to the engraving rules for CWN, 
accidentals outside the KS are not repeated within the current bar. 
This 1-bar unit can therefore be considered the length of the musician’s attention span when reading a piece for performance. 
It is also a common period for changes in local key, 
as is often the case, for example, in leadsheets, with one chord per bar (although, in general, this rule is by no means systematic).
This assumption of dividing the input data into bars is not required in the articles cited above, 
whether they follow a machine learning or an algorithmic approach. 
Instead, some studies use a sliding window of a parameterised length (in terms of the number of notes)
estimated empirically~\cite{Meredith06ps13}. 
In this regard, our procedure is more restrictive.
However, this approach proves useful when dealing with quantised MIDI data, 
particularly during the final stage of a music transcription process, 
after rhythmic quantisation.

The principle behind our algorithm is to estimate, based on input MIDI pitches and bar positions, 
the accidentals that will actually appear in the printed score, in accordance with CWN engraving conventions,
and, at the same time, to deduce information regarding the global Key Signature and local keys 
- we will refer to these as local scales hereafter, since jazz musicians tend not to adhere to a strict tonal framework.  
The initial idea, somewhat naive, is to try to minimize the number of accidentals printed throughout the score, 
leading to several possible choices for global KS and local keys, 
from among several candidates with the modes mentioned above, 
according to the options chosen by the user. 
We then determine the best naming options in each bar, based on the most plausible global KS and local keys.
Shortest-path algorithms are used, at the bar level, to calculate the solution with the lowest cost at each stage.

Our algorithm brings significant improvements
to previous work~\cite{Bouquillard:hal-04458185}.
In addition to extensive architectural and under-the-hood changes, 
the shift from a strictly tonal framework to jazz data management 
required a substantial increase in the number of supported scales, which rose from 30 to 165.
To account for this expansion, 
we propose a generalization of Weber’s distance between tonalities  (a measure originally designed for entirely tonal contexts) 
by extending it to 11 modes spread across 165 distinct scales. 
This extension not only allows us to account for the richer melodic and harmonic vocabulary of jazz, 
but also constitutes a significant contribution that may be of interest to studies in the field of computational musicology.

\smallskip 
In Section~\ref{sec:problem}, we outline the problems addressed, 
and describe Weber's distance between tonalities and our proposed extension. 
Section~\ref{sec:shortest-path} presents a shortest-path algorithm implementing the usual engraving conventions for printing accidentals, and variations used in our procedure. We then detail our algorithm in Section~\ref{sec:method} and present in Section~\ref{sec:evaluation} its evaluation on two kinds of datasets: jazz datasets made of leadsheets, tenor sax soli transcriptions, jazz bass lines 
and traditional tunes on the one hand, and classical scores for piano and monophonic instruments on the other hand.


\section{Names, Scales and Distances}
\label{sec:def}\label{sec:problem}
In this section, 
we describe the problem studied in the paper,
recalling basic notions (Sections~\ref{sec:input}, \ref{sec:output})
and we introduce (Section~\ref{sec:Weber})
a new distance between scales which 
is one key component of our method.

\subsection{Problem Input}\label{sec:input}
Let us assume given in input a  sequence $\nu_1,\ldots, \nu_p$ 
of notes organized in measures (bars),
called a \emph{part}.
It shall typically represent one staff in music notation,
possibly including several voices and chords.
%
Every note $\nu_i$ in the input sequence is defined by: 
\def\inPitch{\mathsf{in}_\mathsf{pit}}
\def\inSimult{\mathsf{in}_\mathsf{sim}}
\def\inBar{\mathsf{in}_\mathsf{bar}}
\begin{description}
\item[$(\inPitch)$] 
a MIDI pitch value in $0..128$, 
\item[$(\inBar)$] 
a boolean flag expressing whether $\nu_i$ belongs to the same bar as $\nu_{i+1}$,
\item[$(\inSimult)$] 
a boolean flag expressing whether  $\nu_i$ and the next note $\nu_{i+1}$ are played simultaneously.
\end{description}

\subsubsection{Spellings.}
The MIDI \emph{pitch} 
of a note $\nu_i$ 
corresponds to the distance in semitones
from a reference lowest note (which has hence a MIDI value 0).
Its value modulo~12, called \emph{pitch class}, is denoted by $\pc(\nu)$.

\begin{figure}
\begin{center}
\includegraphics[width=0.75\textwidth]{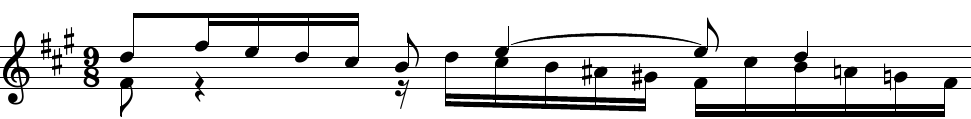}
\end{center}
\caption{J.S. Bach, Fugue in A Major BWV864, measure 33, right hand.} 
\label{fig:BWV864}
\end{figure}

\begin{example} \label{ex:running}
Figure~\ref{fig:BWV864} presents 
the right-hand part in Measure 33 of the Fugue in A Major BWV864
of J.S. Bach.
The MIDI values of the two notes on the first beat of the bar are respectively 
66,  with possible spellings either $\Fs 4$ or $\Gf 4$, 
and 74, 
with possible spellings either
$\Dp 5$, 
or $\Css 5$, 
or $\Eff 5$. 
\end{example}

\subsubsection{Timings.}
We do not assume the {onset} time nor duration of input notes to be given in input.
However, we assume that the notes are enumerated by increasing onset (start time). 
Moreover, we require information on the bar boundaries ($\inBar$)
and note simultaneity ($\inSimult$). 
%
%
We call two notes \emph{simultaneous} when they occur on the same onset, 
because they are involved in the same chord, 
or they belong to ifferent voices and start simultaneously.
However, a \emph{grace-note} $\nu$, 
be it single or involved in 
an ornament (appoggiatura, gruppetto, mordent, trill \etc)
is not considered simultaneous with the next note~$\nu'$, 
but preceding it, 
although in a score, $\nu$ and~$\nu'$ have theoretically the same onset.
By convention, the two flags ($\inSimult$) and ($\inBar$) are set to \textit{false} for the last note~$\nu_p$.

\begin{example}
For instance, the two first notes of Figure~\ref{fig:BWV864} do not really constitute a chord 
but they are {simultaneous} since they share the same onset: 0. 
The time signature for the bar is~9/8, 
hence the two first notes $\Fs$ and $\Dn$ both have a duration of~$\frac{1}{9}$ bar, 
whereas the next semi-quaver $\Fs 5$, 
starting at the onset~$\frac{2}{9}$,
has a duration of~$\frac{1}{18}$.
\end{example}

Providing the above flags  ($\inBar$) and ($\inSimult$) can be done precisely 
only when the time values of the input are quantised,
\ie when then correspond to time position representable in a music score, and expressed in fractions of beats or bars.
Our procedure is therefore especially relevant as a backend task 
in a music transcription framework.

\subsection{Problem Output}\label{sec:output}
Given input notes in the above form,
the goal of the algorithm presented here is to return
the following outcome:
\def\outName{\mathsf{out}_\mathsf{spell}}
\def\outKS{\mathsf{out}_\mathsf{ks}}
\def\outLocal{\mathsf{out}_\mathsf{loc}}
\begin{description} 
\item[$(\outName)$]  \label{out:spell} \label{out:name}
a {spelling} for each note, 
\item[$(\outKS)$]  \label{out:KS}
one estimated global {Key Signature},
\item[$(\outLocal)$]  \label{out:local} \label{out:mode}
one estimated {local scale} for each bar.
\end{description}

\noindent
The \emph{spelling} of a note is made of:
\begin{itemize}
\item a \emph{name} in $\Ap.. \Gp$,
\item a symbol of \emph{accidental}, amongst 
\becarre, \bemol, \doublebemol, \diese, \doublediese, and 
\item an \emph{octave number} in $-2.. 9$.
\end{itemize}

Every note name is associated a unique pitch class: 
0 for~$\Cp$ up to~11 for~$\Bp$, 
and the accidental symbol acts as a pitch class modifier: 
$-2$ for~\doublebemol{},
$-1$ for~\bemol, 
$0$ for~\becarre, 
$+1$ for~\diese, 
and $+2$ for~\doublediese{}.
The accidental~\becarre{} may be omitted in some cases.
\noindent
With the convention that the MIDI pitch~$0$ has spelling $\Cp{-1}$, 
every note spelling can be associated a unique MIDI pitch value.
For instance, $\Gs\,9$ corresponds to the highest MIDI pitch 128,
and the extreme notes~$\Ap 0$ and~$\Cp 7$ of the 88 keys of a piano correspond
to the respective MIDI pitches~21 and~96.
%
%
\begin{figure}
\begin{center}
\begin{tabular}{|r|l|l|l|}
\hline
\ pc\  & spelling$_1$ & spelling$_2$ & spelling$_3$\\
\hline
  0 & \;$\Dff$   & $\Cp$  & \;$\Bs$ \\
  1 & \;$\Df$    & $\Cs$  & [$\Bss$] \\
  2 & \;$\Eff$   & $\Dp$  & \;$\Css$ \\
  3 & [$\Fff$]   & $\Ef$   & \;$\Ds$ \\
  4 & \;$\Ff$    & $\Ep$  & \;$\Dss$ \\
  5 & \;$\Gff$   & $\Fp$  & \;$\Es$ \\
  6 & \;$\Gf$    & $\Fs$  & [$\Ess$] \\
  7 & \;$\Aff$   & $\Gp$  & \;$\Fss$ \\
  8 & \;$\Af$    & $\Gs$  &        \\
  9 & \;$\Bff$   & $\Ap$  & \;$\Gss$ \\
 10 & [$\Cff$] & $\Bf$    & \;$\As$ \\
 11 & \;$\Cf$    & $\Bp$ & \;$\Ass$ \\
 \hline
 \end{tabular}
\end{center}
\caption{Enharmonic spellings for each pitch class.}
\label{fig:enharmonics}
\end{figure}
%
\noindent
The opposite is not however true: 
there exists several (2 or 3) alternative valid spellings for every MIDI value, 
summarised in Figure~\ref{fig:enharmonics} for the 12 pitch classes.
For instance, $\Bs{-2}$, $\Cn{-1}$ and $\Dp\doublebemol{-1}$
are alternative spellings for the MIDI pitch 0.
%
The above objective~$(\outName)$ therefore amounts to chose, 
for each given note $\nu_i$, one of the 2 or 3 alternative spellings 
-- actually, choosing one note name, since, for a given the MIDI pitch of a note $\nu_i$, 
the other elements of a spelling can be obtained from the name.
%
%

A \emph{Key Signature} (KS) is denoted by an integer $k$ 
between
$-7$ and $7$, 
which indicates that, by default, 
$|k|$ note names shall be altered 
by a~\diese{}, 
when $k$ is positive, 
or by a~\bemol{}, 
when $k$ is negative. 
The names of the notes altered are defined according to the order of fifths: 
$\Fs, \Cs, \Gs, \Ds, \As, \Es, \Bs$ for $k>0$,
and 
$\Bf, \Ef, \Af, \Df, \Gf, \Cf, \Ff,$ for $k<0$.
%
%
\newcommand{\largebecarre}{\ \phantom{\diese} \ }
\newcommand{\largediese}{\ \diese\ \ }
\newcommand{\largebemol}{\ \bemol\ \ }
\begin{figure*}
\begin{tabular}{c|lll|ccccccccc|ccc|}
{\small Key Signature} & $-7$ & $-6$ & $-5$ & $-4$ & $-3$ & $-2$ & $-1$ & 
$0$ & $1$ & $2$ & $3$ & $4$ & $5$ & $6$ & $7$ \\
\hline
{\small major scale} & 
\;$\Cf$ & \;$\Gf$ & \;$\Df$ & \;$\Af$ & \;$\Ef$ & \;$\Bf$ & \;$\Fp$ & 
\;$\Cp$ & \;$\Gp$ & \;$\Dp$ & \;$\Ap$ & \;$\Ep$ & \;$\Bp$ & \;$\Fs$ & \;$\Cs$\\
{\small minor scales} & 
\;$\Af$ & \;$\Ef$ & \;$\Bf$ & \;$\Fp$ & \;$\Cp$ & \;$\Gp$ & \;$\Dp$ & 
\;$\Ap$ & \;$\Ep$ & \;$\Bp$ & \;$\Fs$ & \;$\Cs$ & \;$\Gs$ & \;$\Ds$ & \;$\As$\\[2pt]
\begin{tabular}{|c|}
\hline
$\Cp$\\
\hline
$\Dp$\\ 
\hline
$\Ep$\\ 
\hline
$\Fp$\\ 
\hline
$\Gp$\\ 
\hline
$\Ap$\\ 
\hline
$\Bp$\\ 
\hline
\end{tabular}
& 
\begin{tabular}{|c|}
\hline
\largebemol \\
\hline
\largebemol\\ 
\hline
\largebemol\\ 
\hline
\largebemol\\ 
\hline
\largebemol\\ 
\hline
\largebemol\\ 
\hline
\largebemol\\ 
\hline
\end{tabular} 
& 
\begin{tabular}{|c|}
\hline
\largebemol\\
\hline
\largebemol\\ 
\hline
\largebemol\\ 
\hline
\largebecarre\\ 
\hline
\largebemol\\ 
\hline
\largebemol\\ 
\hline
\largebemol\\ 
\hline
\end{tabular} 
& 
\begin{tabular}{|c|}
\hline
\largebecarre\\
\hline
\largebemol\\ 
\hline
\largebemol\\ 
\hline
\largebecarre\\ 
\hline
\largebemol\\ 
\hline
\largebemol\\ 
\hline
\largebemol\\ 
\hline
\end{tabular} 
& 
\begin{tabular}{|c|}
\hline
\largebecarre\\
\hline
\largebemol\\ 
\hline
\largebemol\\ 
\hline
\largebecarre\\ 
\hline
\largebecarre\\ 
\hline
\largebemol\\ 
\hline
\largebemol\\ 
\hline
\end{tabular} 
& 
\begin{tabular}{|c|}
\hline
\largebecarre\\
\hline
\largebecarre\\ 
\hline
\largebemol\\ 
\hline
\largebecarre\\ 
\hline
\largebecarre\\ 
\hline
\largebemol\\ 
\hline
\largebemol\\ 
\hline
\end{tabular} 
& 
\begin{tabular}{|c|}
\hline
\largebecarre\\
\hline
\largebecarre\\ 
\hline
\largebemol\\ 
\hline
\largebecarre\\ 
\hline
\largebecarre\\ 
\hline
\largebecarre\\ 
\hline
\largebemol\\ 
\hline
\end{tabular} 
& 
\begin{tabular}{|c|}
\hline
\largebecarre\\
\hline
\largebecarre\\ 
\hline
\largebecarre\\ 
\hline
\largebecarre\\ 
\hline
\largebecarre\\ 
\hline
\largebecarre\\ 
\hline
\largebemol\\ 
\hline
\end{tabular} 
& 
\begin{tabular}{|c|}
\hline
\largebecarre\\
\hline
\largebecarre\\ 
\hline
\largebecarre\\ 
\hline
\largebecarre\\ 
\hline
\largebecarre\\ 
\hline
\largebecarre\\ 
\hline
\largebecarre\\ 
\hline
\end{tabular} 
& 
\begin{tabular}{|c|}
\hline
\largebecarre\\
\hline
\largebecarre\\ 
\hline
\largebecarre\\ 
\hline
\largediese\\ 
\hline
\largebecarre\\ 
\hline
\largebecarre\\ 
\hline
\largebecarre\\ 
\hline
\end{tabular} 
& 
\begin{tabular}{|c|}
\hline
\largediese\\
\hline
\largebecarre\\ 
\hline
\largebecarre\\ 
\hline
\largediese\\ 
\hline
\largebecarre\\ 
\hline
\largebecarre\\ 
\hline
\largebecarre\\ 
\hline
\end{tabular} 
& 
\begin{tabular}{|c|}
\hline
\largediese\\
\hline
\largebecarre\\ 
\hline
\largebecarre\\ 
\hline
\largediese\\ 
\hline
\largediese\\ 
\hline
\largebecarre\\ 
\hline
\largebecarre\\ 
\hline
\end{tabular} 
& 
\begin{tabular}{|c|}
\hline
\largediese\\
\hline
\largediese\\ 
\hline
\largebecarre\\ 
\hline
\largediese\\ 
\hline
\largediese\\ 
\hline
\largebecarre\\ 
\hline
\largebecarre\\ 
\hline
\end{tabular} 
& 
\begin{tabular}{|c|}
\hline
\largediese\\
\hline
\largediese\\ 
\hline
\largebecarre\\ 
\hline
\largediese\\ 
\hline
\largediese\\ 
\hline
\largediese\\ 
\hline
\largebecarre\\ 
\hline
\end{tabular} 
& 
\begin{tabular}{|c|}
\hline
\largediese\\
\hline
\largediese\\ 
\hline
\largediese\\ 
\hline
\largediese\\ 
\hline
\largediese\\ 
\hline
\largediese\\ 
\hline
\largebecarre\\ 
\hline
\end{tabular} 
& 
\begin{tabular}{|c|}
\hline
\largediese\\
\hline
\largediese\\ 
\hline
\largediese\\ 
\hline
\largediese\\ 
\hline
\largediese\\ 
\hline
\largediese\\ 
\hline
\largediese\\ 
\hline
\end{tabular} 
\end{tabular}
\caption{Key Signatures and tonic of usual scales.}
\label{fig:tons}\label{fig:keys}
\end{figure*}
%
In Figure~\ref{fig:keys}, 
we describe 15 Key Signatures and the tonic of associated major and minor scales.
In theory, 
the list of Key Signatures can be extended on the right and on the left, 
respectively through double sharps and double flats.
For instance, with $k=8$ ($\Gs$ major), 
$\Fp$ is altered with a double sharps ($\Fss$), 
with $k=9$ ($\Ds$ major), 
$\Fp$ and $\Cp$ are altered with double sharps ($\Fss$, $\Css$), \etc. 
We do not consider the case of extended KS in this work, 
as they are very rarely found.

In a score, a KS indication is placed at the beginning of a part. 
The KS can be changed during a part (at the beginning of a bar).
From the notational point of view (\ie for engraving), 
the choice of a KS will influence drastically 
the spelling of notes in every bar,
and the display of accidentals, 
hence the readability of the score, 
as recalled in Section~\ref{sec:convention}.
	
A \emph{mode} is a sequence of intervals.
In this work, we consider in the following 9 heptatonic modes, 
as well as two blues modes: 
\begin{itemize}
\item ionian (or major mode), 
\item dorian ($\Dp$ mode),
\item phrygian ($\Ep$ mode),
\item lydian ($\Fp$ mode),
\item mixolydian ($\Gp$ mode),
\item aeolian ($\Ap$ mode or natural minor), 
\item melodic minor, 
\item harmonic minor, 
\item locrian ($\Bp$ mode), 
\item major blues,
\item minor blues.
\end{itemize}

We call \emph{scale} 
the pairing of a KS and a mode. 
It is functionally identical to the usual definition of a mode anchored to a tonic. 
%
In a tonal context,  "scale" and "key" are synonymous concepts.
From a musical point of view, 
the key of a piece identifies a diatonic scale, 
whose first note (amongst seven notes), 
called the \emph{tonic} note, 
represents the main tonal focus of the piece.

Some scales can induce accidentals outside the KS, 
that we call here \emph{characteristic accidentals}.
It is for example the case of the leading tone at the seventh degree of 
harmonic minor scales.
Intuitively, these accidentals, when printed, help to recognise the scale at sight.


In both major and minor scales,
the keys associated with Key Signatures 
respectively 
-7 and 5, 
-5 and 7, and 
-6 and 6 
have tonics with the same pitch class, but different names.
These keys are called \emph{enharmonic}.
They correspond to the 3 first and 3 last columns in Figure~\ref{fig:keys}.
Melodies written in either of two enharmonic keys
cannot be distinguished by ear (in equal temperaments) as changing a key for its enharmonic preserves not only the intervals but also the pitch class of every note.
%
Therefore, spelling in one or the other of enharmonic keys 
is essentially a matter of choice.
Usually, the keys with KS~5 (5~sharps, \eg $\Bp$ major) 
or KS~{-5} (5~flats, \eg $\Df$ major) are preferred
over their enharmonic equivalents
with respectively 
KS~{-7} (\eg $\Cf$ major) 
and KS~7 (\eg $\Cs$ major).
But it is not always the case, 
for instance the 
Prelude and Fugue, BWV~848,
of Bach's Well-Tempered Clavier
are in $\Cs$ major. 

\subsection{Distances between Scales}	
\label{sec:Weber}
Gottfried Weber defines~\cite{Weber}
a measure of distance between major and harmonic minor scales, 
which has been used for MIR tasks related to key estimation~\cite{Feisthauer20smc}.
%
\def\AP{\textsf{A}}
\def\BP{\textsf{B}}
\def\CP{\textsf{C}}
\def\DP{\textsf{D}}
\def\EP{\textsf{E}}
\def\FP{\textsf{F}}
\def\GP{\textsf{G}}
\def\aP{\textsf{a}}
\def\bP{\textsf{b}}
\def\cP{\textsf{c}}
\def\dP{\textsf{d}}
\def\eP{\textsf{e}}
\def\fP{\textsf{f}}
\def\gP{\textsf{g}}
\def\AS{\textsf{A}\musSharp} 
\def\BS{\textsf{B}\musSharp}
\def\CS{\textsf{C}\musSharp}
\def\DS{\textsf{D}\musSharp}
\def\ES{\textsf{E}\musSharp}
\def\FS{\textsf{F}\musSharp}
\def\GS{\textsf{G}\musSharp}
\def\aS{\textsf{a}\musSharp} 
\def\bS{\textsf{b}\musSharp}
\def\cS{\textsf{c}\musSharp}
\def\dS{\textsf{d}\musSharp}
\def\eS{\textsf{e}\musSharp}
\def\fS{\textsf{f}\musSharp}
\def\gS{\textsf{g}\musSharp}
\def\AF{\textsf{A}\musFlat} 
\def\BF{\textsf{B}\musFlat}
\def\CF{\textsf{C}\musFlat}
\def\DF{\textsf{D}\musFlat}
\def\EF{\textsf{E}\musFlat}
\def\FF{\textsf{F}\musFlat}
\def\GF{\textsf{G}\musFlat}
\def\aF{\textsf{a}\musFlat} 
\def\bF{\textsf{b}\musFlat}
\def\cF{\textsf{c}\musFlat}
\def\dF{\textsf{d}\musFlat}
\def\eF{\textsf{e}\musFlat}
\def\fF{\textsf{f}\musFlat}
\def\gF{\textsf{g}\musFlat}
\begin{figure}
\begin{center}
\scalebox{0.80}{%
\rotatebox{0}{%
\begin{tabular}{|r|l|rrrrrrrrrrrrrrrrrrrrrrrrrrrrrr|}
\hline
KS  &   & -7 & -6 & -5 & -4 & -3 & -2 & -1  &  0 &  1 &  2 &  3 &  4 &  5 &  6 &  7 & -7 & -6 & -5 & -4 & -3 & -2 & -1 &  0 &  1 &  2 &  3 &  4 &  5 &  6 &  7 \\
    &   & \CF & \GF & \DF & \AF & \EF & \BF & \FP & \CP & \GP & \DP & \AP & \EP & \BP & \FS & \CS & \aF & \eF & \bF & \fP & \cP & \gP & \dP & \aP & \eP & \bP & \fS & \cS & \gS & \dS & \aS \\
\hline
-7 & \CF &  0 &  1 &  2 &  2 &  3 &  4 &  4 &  5 &  6 &  6 &  7 &  8 &  8 &  9 & 10 &  1 &  2 &  3 &  3 &  4 &  5 &  5 &  6 &  7 &  7 &  8 &  9 &  9 & 10 & 11 \\
-6 & \GF &  1 &  0 &  1 &  2 &  2 &  3 &  4 &  4 &  5 &  6 &  6 &  7 &  8 &  8 &  9 &  2 &  1 &  2 &  3 &  3 &  4 &  5 &  5 &  6 &  7 &  7 &  8 &  9 &  9 & 10 \\
-5 & \DF &  2 &  1 &  0 &  1 &  2 &  2 &  3 &  4 &  4 &  5 &  6 &  6 &  7 &  8 &  8 &  2 &  2 &  1 &  2 &  3 &  3 &  4 &  5 &  5 &  6 &  7 &  7 &  8 &  9 &  9 \\
-4 & \AF &  2 &  2 &  1 &  0 &  1 &  2 &  2 &  3 &  4 &  4 &  5 &  6 &  6 &  7 &  8 &  1 &  2 &  2 &  1 &  2 &  3 &  3 &  4 &  5 &  5 &  6 &  7 &  7 &  8 &  9 \\
-3 & \EF &  3 &  2 &  2 &  1 &  0 &  1 &  2 &  2 &  3 &  4 &  4 &  5 &  6 &  6 &  7 &  2 &  1 &  2 &  2 &  1 &  2 &  3 &  3 &  4 &  5 &  5 &  6 &  7 &  7 &  8 \\
-2 & \BF &  4 &  3 &  2 &  2 &  1 &  0 &  1 &  2 &  2 &  3 &  4 &  4 &  5 &  6 &  6 &  3 &  2 &  1 &  2 &  2 &  1 &  2 &  3 &  3 &  4 &  5 &  5 &  6 &  7 &  7 \\
-1 & \FP &  4 &  4 &  3 &  2 &  2 &  1 &  0 &  1 &  2 &  2 &  3 &  4 &  4 &  5 &  6 &  3 &  3 &  2 &  1 &  2 &  2 &  1 &  2 &  3 &  3 &  4 &  5 &  5 &  6 &  7 \\
 0 & \CP &  5 &  4 &  4 &  3 &  2 &  2 &  1 &  0 &  1 &  2 &  2 &  3 &  4 &  4 &  5 &  4 &  3 &  3 &  2 &  1 &  2 &  2 &  1 &  2 &  3 &  3 &  4 &  5 &  5 &  6 \\
 1 & \GP &  6 &  5 &  4 &  4 &  3 &  2 &  2 &  1 &  0 &  1 &  2 &  2 &  3 &  4 &  4 &  5 &  4 &  3 &  3 &  2 &  1 &  2 &  2 &  1 &  2 &  3 &  3 &  4 &  5 &  5 \\
 2 & \DP &  6 &  6 &  5 &  4 &  4 &  3 &  2 &  2 &  1 &  0 &  1 &  2 &  2 &  3 &  4 &  5 &  5 &  4 &  3 &  3 &  2 &  1 &  2 &  2 &  1 &  2 &  3 &  3 &  4 &  5 \\
 3 & \AP &  7 &  6 &  6 &  5 &  4 &  4 &  3 &  2 &  2 &  1 &  0 &  1 &  2 &  2 &  3 &  6 &  5 &  5 &  4 &  3 &  3 &  2 &  1 &  2 &  2 &  1 &  2 &  3 &  3 &  4 \\
 4 & \EP &  8 &  7 &  6 &  6 &  5 &  4 &  4 &  3 &  2 &  2 &  1 &  0 &  1 &  2 &  2 &  7 &  6 &  5 &  5 &  4 &  3 &  3 &  2 &  1 &  2 &  2 &  1 &  2 &  3 &  3 \\
 5 & \BP &  8 &  8 &  7 &  6 &  6 &  5 &  4 &  4 &  3 &  2 &  2 &  1 &  0 &  1 &  2 &  7 &  7 &  6 &  5 &  5 &  4 &  3 &  3 &  2 &  1 &  2 &  2 &  1 &  2 &  3 \\
 6 & \FS &  9 &  8 &  8 &  7 &  6 &  6 &  5 &  4 &  4 &  3 &  2 &  2 &  1 &  0 &  1 &  8 &  7 &  7 &  6 &  5 &  5 &  4 &  3 &  3 &  2 &  1 &  2 &  2 &  1 &  2 \\
 7 & \CS & 10 &  9 &  8 &  8 &  7 &  6 &  6 &  5 &  4 &  4 &  3 &  2 &  2 &  1 &  0 &  9 &  8 &  7 &  7 &  6 &  5 &  5 &  4 &  3 &  3 &  2 &  1 &  2 &  2 &  1 \\
-7 & \aF &  1 &  2 &  2 &  1 &  2 &  3 &  3 &  4 &  5 &  5 &  6 &  7 &  7 &  8 &  9 &  0 &  1 &  2 &  2 &  3 &  4 &  4 &  5 &  6 &  6 &  7 &  8 &  8 &  9 & 10 \\
-6 & \eF &  2 &  1 &  2 &  2 &  1 &  2 &  3 &  3 &  4 &  5 &  5 &  6 &  7 &  7 &  8 &  1 &  0 &  1 &  2 &  2 &  3 &  4 &  4 &  5 &  6 &  6 &  7 &  8 &  8 &  9 \\
-5 & \bP &  3 &  2 &  1 &  2 &  2 &  1 &  2 &  3 &  3 &  4 &  5 &  5 &  6 &  7 &  7 &  2 &  1 &  0 &  1 &  2 &  2 &  3 &  4 &  4 &  5 &  6 &  6 &  7 &  8 &  8 \\
-4 & \fP &  3 &  3 &  2 &  1 &  2 &  2 &  1 &  2 &  3 &  3 &  4 &  5 &  5 &  6 &  7 &  2 &  2 &  1 &  0 &  1 &  2 &  2 &  3 &  4 &  4 &  5 &  6 &  6 &  7 &  8 \\
-3 & \cP &  4 &  3 &  3 &  2 &  1 &  2 &  2 &  1 &  2 &  3 &  3 &  4 &  5 &  5 &  6 &  3 &  2 &  2 &  1 &  0 &  1 &  2 &  2 &  3 &  4 &  4 &  5 &  6 &  6 &  7 \\
-2 & \gP &  5 &  4 &  3 &  3 &  2 &  1 &  2 &  2 &  1 &  2 &  3 &  3 &  4 &  5 &  5 &  4 &  3 &  2 &  2 &  1 &  0 &  1 &  2 &  2 &  3 &  4 &  4 &  5 &  6 &  6 \\
-1 & \dP &  5 &  5 &  4 &  3 &  3 &  2 &  1 &  2 &  2 &  1 &  2 &  3 &  3 &  4 &  5 &  4 &  4 &  3 &  2 &  2 &  1 &  0 &  1 &  2 &  2 &  3 &  4 &  4 &  5 &  6 \\
 0 & \aP &  6 &  5 &  5 &  4 &  3 &  3 &  2 &  1 &  2 &  2 &  1 &  2 &  3 &  3 &  4 &  5 &  4 &  4 &  3 &  2 &  2 &  1 &  0 &  1 &  2 &  2 &  3 &  4 &  4 &  5 \\
 1 & \eP &  7 &  6 &  5 &  5 &  4 &  3 &  3 &  2 &  1 &  2 &  2 &  1 &  2 &  3 &  3 &  6 &  5 &  4 &  4 &  3 &  2 &  2 &  1 &  0 &  1 &  2 &  2 &  3 &  4 &  4 \\
 2 & \bP &  7 &  7 &  6 &  5 &  5 &  4 &  3 &  3 &  2 &  1 &  2 &  2 &  1 &  2 &  3 &  6 &  6 &  5 &  4 &  4 &  3 &  2 &  2 &  1 &  0 &  1 &  2 &  2 &  3 &  4 \\
 3 & \fS &  8 &  7 &  7 &  6 &  5 &  5 &  4 &  3 &  3 &  2 &  1 &  2 &  2 &  1 &  2 &  7 &  6 &  6 &  5 &  4 &  4 &  3 &  2 &  2 &  1 &  0 &  1 &  2 &  2 &  3 \\
 4 & \cS &  9 &  8 &  7 &  7 &  6 &  5 &  5 &  4 &  3 &  3 &  2 &  1 &  2 &  2 &  1 &  8 &  7 &  6 &  6 &  5 &  4 &  4 &  3 &  2 &  2 &  1 &  0 &  1 &  2 &  2 \\
 5 & \gS &  9 &  9 &  8 &  7 &  7 &  6 &  5 &  5 &  4 &  3 &  3 &  2 &  1 &  2 &  2 &  8 &  8 &  7 &  6 &  6 &  5 &  4 &  4 &  3 &  2 &  2 &  1 &  0 &  1 &  2 \\
 6 & \dS & 10 &  9 &  9 &  8 &  7 &  7 &  6 &  5 &  5 &  4 &  3 &  3 &  2 &  1 &  2 &  9 &  8 &  8 &  7 &  6 &  6 &  5 &  4 &  4 &  3 &  2 &  2 &  1 &  0 &  1 \\
 7 & \aS & 11 & 10 &  9 &  9 &  8 &  7 &  7 &  6 &  5 &  5 &  4 &  3 &  3 &  2 &  1 & 10 &  9 &  8 &  8 &  7 &  6 &  6 &  5 &  4 &  4 &  3 &  2 &  2 &  1 &  0 \\
\hline
\end{tabular}
} 
} 
\end{center}
\caption{The table of relationship between keys, by G. Weber~\cite{Weber}}
Keys in major scale are uppercase, keys in harmonic minor scale are lowercase.
\label{fig:Weber}
\end{figure}
%
The Weber distance between two keys is 
the length of a shortest path between them in a 2D grid
(Figure~\ref{fig:Weber}).
Each node in the grid 
represents a specific key~$K$ with 4 neighbors 
that are considered close to~$K$, 
either because they differ from~$K$ by only one note 
(dominant key, subdominant key, relative key) 
or because they have the same tonic 
(homonym keys: same tonic but different mode).

\begin{figure}[t]

\def\major{\mathsf{maj}}
\def\minorharm{\mathsf{min}}
\def\minormel{\mathsf{mel}}
\def\ionian{\mathsf{io}}
\def\dorian{\mathsf{do}}
\def\phrygian{\mathsf{ph}}
\def\lydian{\ell\mathsf{y}}
\def\mixolydian{\mathsf{mx}}
\def\aeolian{\mathsf{ae}}
\def\locrian{\ell\mathsf{o}}
                   
\centering 
\begin{tikzpicture}[x=(15:.5cm), y=(90:.5cm), z=(330:.5cm), >=stealth]
\draw  [dashed] (0, 0, 0) -- (0, 0, 12);
\draw  [dashed] (3, 0, 0) -- (3, 0, 12);
\foreach \z in {0, 4, 8, 12} \foreach \x in {0,1,2}
  \foreach \y [evaluate={\b=random(0, 9);}] in {0,1,2}
    \filldraw [fill=white] (\x, \y, \z) -- (\x+1, \y, \z) -- (\x+1, \y+1, \z) --
      (\x, \y+1, \z) -- cycle; 
\foreach \z in {0, 4, 8, 12} \foreach \x in {0,1,2,3}
    \draw (\x, 0, \z) -- (\x, -0.3, \z) (\x, 3, \z) -- (\x, 3.3, \z);	
\foreach \z in {0, 4, 8, 12} \foreach \y in {0,1,2,3}
    \draw (0, \y, \z) -- (-0.3, \y, \z) (3, \y, \z) -- (3.3, \y, \z);	
    
\node at (0.5, 0.5, 0)  [yslant=tan(15)] {$\scriptstyle\Cp^{\dorian}$};
\node at (1.5, 0.5, 0)  [yslant=tan(15)] {$\scriptstyle\Cp^{\lydian}$};
\node at (2.5, 0.5, 0)  [yslant=tan(15)] {$\scriptstyle\Ap^{\dorian}$};
\node at (0.5, 1.5, 0)  [yslant=tan(15)] {$\scriptstyle\Fp^{\dorian}$};
\node at (1.5, 1.5, 0)  [yslant=tan(15)] {$\scriptstyle\Fp^{\lydian}$};
\node at (2.5, 1.5, 0)  [yslant=tan(15)] {$\scriptstyle\Dp^{\dorian}$};
\node at (0.5, 2.5, 0)  [yslant=tan(15)] {$\scriptstyle\Bf^{\dorian}$};
\node at (1.5, 2.5, 0)  [yslant=tan(15)] {$\scriptstyle\Bf^{\lydian}$};
\node at (2.5, 2.5, 0)  [yslant=tan(15)] {$\scriptstyle\Gp^{\dorian}$};

\node at (0.5, 0.5, 2)  [yslant=tan(15)] {$\scriptstyle\mbox{Blues}$};

\node at (0.5, 0.5, 4)  [yslant=tan(15)] {$\scriptstyle\Gp^{\aeolian}$};
\node at (1.5, 0.5, 4)  [yslant=tan(15)] {$\scriptstyle\Gp^{\ionian}$};
\node at (2.5, 0.5, 4)  [yslant=tan(15)] {$\scriptstyle\Ep^{\aeolian}$};
\node at (0.5, 1.5, 4)  [yslant=tan(15)] {$\scriptstyle\Cp^{\aeolian}$};
\node at (1.5, 1.5, 4)  [yslant=tan(15)] {$\scriptstyle\Cp^{\ionian}$};
\node at (2.5, 1.5, 4)  [yslant=tan(15)] {$\scriptstyle\Ap^{\aeolian}$};
\node at (0.5, 2.5, 4)  [yslant=tan(15)] {$\scriptstyle\Fp^{\aeolian}$};
\node at (1.5, 2.5, 4)  [yslant=tan(15)] {$\scriptstyle\Fp^{\ionian}$};
\node at (2.5, 2.5, 4)  [yslant=tan(15)] {$\scriptstyle\Dp^{\aeolian}$};

\node at (0.5, 0.5, 8)  [yslant=tan(15)] {$\scriptstyle\Dp^{\phrygian}$};
\node at (1.5, 0.5, 8)  [yslant=tan(15)] {$\scriptstyle\Dp^{\mixolydian}$};
\node at (2.5, 0.5, 8)  [yslant=tan(15)] {$\scriptstyle\Bp^{\phrygian}$};
\node at (0.5, 1.5, 8)  [yslant=tan(15)] {$\scriptstyle\Gp^{\phrygian}$};
\node at (1.5, 1.5, 8)  [yslant=tan(15)] {$\scriptstyle\Gp^{\mixolydian}$};
\node at (2.5, 1.5, 8)  [yslant=tan(15)] {$\scriptstyle\Ep^{\phrygian}$};
\node at (0.5, 2.5, 8)  [yslant=tan(15)] {$\scriptstyle\Cp^{\phrygian}$};
\node at (1.5, 2.5, 8)  [yslant=tan(15)] {$\scriptstyle\Cp^{\mixolydian}$};
\node at (2.5, 2.5, 8)  [yslant=tan(15)] {$\scriptstyle\Ap^{\phrygian}$};

\node at (0.5, 0.5, 12)  [yslant=tan(15)] {$\scriptstyle\Ap^{\locrian}$};
\node at (1.5, 0.5, 12)  [yslant=tan(15)] {$\scriptstyle\Ap^{\dorian}$};
\node at (2.5, 0.5, 12)  [yslant=tan(15)] {$\scriptstyle\Fs^{\locrian}$};
\node at (0.5, 1.5, 12)  [yslant=tan(15)] {$\scriptstyle\Dp^{\locrian}$};
\node at (1.5, 1.5, 12)  [yslant=tan(15)] {$\scriptstyle\Dp^{\dorian}$};
\node at (2.5, 1.5, 12)  [yslant=tan(15)] {$\scriptstyle\Bp^{\locrian}$};
\node at (0.5, 2.5, 12)  [yslant=tan(15)] {$\scriptstyle\Gp^{\locrian}$};
\node at (1.5, 2.5, 12)  [yslant=tan(15)] {$\scriptstyle\Gp^{\dorian}$};
\node at (2.5, 2.5, 12)  [yslant=tan(15)] {$\scriptstyle\Ep^{\locrian}$};

\draw [dashed] (3, 3, 0) -- (3, 3, 12);
\end{tikzpicture}%
\caption{Weber distance generalised to common jazz modes.\label{fig:distance}} 
\label{fig:Weber2}
\label{fig:Weberplus}
\end{figure}
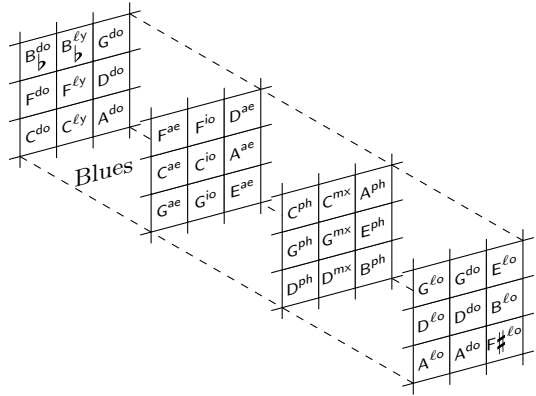

We propose in this work an extension of Weber's distance 
to the seven modern diatonic modes (ionian, dorian, \etc), melodic and harmonic minor modes, and two blues modes, 
used for the estimation of local scales in the step described in Section~\ref{sec:modal}
of the PS algorithm. 
%
We use for this purpose a 3D structure 
represented in Figure~\ref{fig:Weberplus}. 
The modes are grouped by pairs, 
following the relationship between a major (\ie ionian) scale and its minor (\ie aeolian = 
natural minor) relative. 
It means in particular that a descending minor third always separates the first note of the "major-like" scale 
from the first note of its "minor-like" counterpart; 
for example, $\Fp$~lydian has~$\Dp$~dorian as its "modal relative". 
From every pair of "relative" modes we derive a new 2D grid similar to the original grid of~\cite{Weber}. 
%
All of the grids are aligned such that scales with a given nature of third (major or minor) 
and containing exactly the same notes are always placed on the same line; 
for example, the $\Fp$~lydian scale from the lydian-dorian grid,  with a major third,   
is behind $\Cp$~ionian from the ionian-aeolian grid 
which is itself behind $\Gp$ mixolydian from the mixolydian-phrygian grid.
 
We add to this 3D structure 
the two other minor modes 
(melodic and harmonic), 
in the same places as their aeolian counterparts (with identical tonics), 
and finally the major and minor blues scales.
The latter constitute 
a supplementary 2D grid 
placed between the lydian-dorian and ionian-aeolian grids,
since the lydian distinctive augmented fourth is also present in the minor blues mode 
and the dorian mode is almost entirely contained in the major as well as in the minor blues mode.

We assign a cost of 1 to any move in a straight line from one 2D grid to an immediate neighbouring 2D grid, and to any horizontal or vertical move within a single 2D grid. 
\noindent
Finally, the \emph{distance} between two scales 
is the minimum number of moves in the 3D grid to go from one to the other. 

\section{Shortest Path view of  Accidentals' Engraving}
\label{sec:shortest-path}
In this section we recall the conventions for printing accidentals in music scores, 
present a shortest-path algorithm for ensuring them (Section~\ref{sec:engraving}), 
and some variations (Sections~\ref{sec:initialstate}, \ref{sec:chords})
of this algorithm that we are using in our 
procedure.

\subsection{Conventional Engraving of Accidentals}
\label{sec:convention} \label{sec:engraving}
For readability reasons, 
some accidentals are not printed in engraved scores.
Following a principle of parsimony, 
the notational conventions~\cite{Gould11Notation} are roughly as follows:
\begin{quote}
\it accidentals already in the Key Signature are omitted by default, 
and other accidentals need not be repeated in the same bar.
\end{quote}
There is an additional restriction to this rule, which we will treat 
as an option in the following sections:

\def\optOctave{\mathsf{opt}_\mathsf{oct}}
\begin{quote}
$(\optOctave)$~{\it An accidental applies only to the pitch at which it is written:
each additional octave for the same pitch class requires a further accidental}~\cite{Gould11Notation}.
\end{quote}

\begin{example} \label{ex:running-spell}
In Figure~\ref{fig:BWV864} (see also Example~\ref{ex:running}), 
the chosen spelling of the two notes on the first beat
does not induce any printed accidental, 
as $\Fs$ and $\Dn$ are included in the Key Signature.
An alternative spelling for these two notes could be 
$\Gf$, 
$\Dn$ 
but it would generate an additional accidental 
since the signature does not include $\Gf$.
\end{example}

\smallskip\noindent
In order to ensure the above principle, 
we consider a \emph{state}, which is a mapping of note names,  
and octaves in the case of $(\optOctave)$,
into accidental symbols.
%
Given a global scale $S$, 
we start a bar in an initial state $\sigma_0$ 
containing exactly the accidentals defined by the KS~of~$S$, as recalled in Figure~\ref{fig:keys}.
All notes in the bar are then processed to determine which accidentals are printed.
Assume that we are in a state $\sigma$, 
and want to process a note $\nu$ with pitch class $p$.
There are up to three possible choices of name~$e$ and accidental~$a$ for~$p$, 
according to the table in Figure~\ref{fig:enharmonics}, 
and each defines a transition to a new state $\sigma'$ as follows:
\begin{itemize}
\item
if $\sigma(e) = a$, then $\sigma' = \sigma$,
and the accidental $a$ is not printed for $\nu$, 
\item
otherwise, 
$\sigma'(e) = a\neq\sigma(e)$ ($\sigma'$ is identical to $\sigma$ for the other names), 
and $a$ is printed for $\nu$.
\end{itemize}

\begin{example}
In Figure~\ref{fig:BWV864}, 
for instance, before onset~$\frac{5}{9}$, the spelling state is composed 
of $\Fs$, $\Cs$, $\Gs$ 
(\ie $\sigma(\Fp) = \sigma(\Cp)  = \sigma(\Gp) = \diese$)
and every other note name is mapped to $\becarre$. 
At onset $\frac{5}{9}$ however, the state changes for the first time in the bar 
with an update of~$\An$ into~$\As$. 
The next onset also induces a change in the state with $\Gn$ being replaced by $\Gs$, 
which is a note present in the ascending minor melodic mode of $\Bp$. 
It is interesting to note that Bach used it in a descending motion, 
therefore a Pitch Spelling process relying too much on motion direction between notes 
would have failed here. 
The spelling state then remains the same until onset $\frac{15}{18}$ 
with the return of $\An$ 
(belonging to the natural minor mode of $\Bp$) 
and finally $\Gn$ at onset $\frac{16}{18}$, 
hence the last state of the bar is the same as the one it started with.
\end{example}

By assigning an integral \emph{cost value} to each transition, 
related to the number of printed accidentals, 
we can compute an optimal path for each bar and scale~$S$
(\ie a path minimizing the number of printed accidentals).
For this purpose, we can use a 
Viterbi algorithm~\cite{Huang08advanceddynamic},
tagging every state $\sigma$ reachable after processing the~$n$ first notes of the bar, 
with the cumulated cost of the best path from 
$\sigma_0$ into $\sigma$.
The time complexity is linear in the number of states plus the number of transitions.
In the worst case, the number of states reachable after reading $n$ notes
can be exponential in $n$, 
hence the algorithm is exponential in the number of notes in the bar.
However, we can prune unnecessary search branches by 
building the reachable states and transitions on-the-fly, which keeps computation time reasonable in practice.

We were vague above regarding the definition of the cost values associated to transitions.
Actually they are in the literature, to our knowledge, no precise definition of an optimal spelling following the above rule.
We consider below two options for counting accidentals in a transition from $\sigma$ to $\sigma'$ reading $\nu$:
\def\optAccidCount{\mathsf{opt}_\mathsf{ac}}
\def\optAccidDist{\mathsf{opt}_\mathsf{ad}}
\begin{description}
\item[$(\optAccidCount)$] 0 if $a$ is not printed, $1$ if $a$ is printed and single (\bemol, \becarre{} or \diese),\\
$2$ if $a$ is printed and double (\doublebemol{} or \doublediese{}).
\item[$(\optAccidDist)$] the distance between $\sigma(e)$ to $\sigma'(e)$, 
considering the position of accidentals on a line: \doublebemol, \bemol, \becarre, \diese, \doublediese.
\end{description}
Both options will enforce printing as few accidentals as possible, with different focus. 
The $(\optAccidCount)$ option heavily penalizes double accidentals, 
which are generally quite rare in notation, 
but can be useful though for avoiding too many changes.
The $(\optAccidDist)$ option penalizes changes in the color of accidental (from \bemol{} to \becarre{} or vice versa).

\subsection{Extensions to scales}
\label{sec:initialstate}
The best-path algorithm of Section~\ref{sec:engraving}
considers only the KS of a scale $S$ for the definition of the initial state,
complying to engraving conventions for printing or not accidentals.
However, in the following, we shall use best paths (and their cost) not only for PS but also 
for the estimation of global and local scales (and not only KS). 
In this context, it is important to be able to distinguish between scales with the same KS but different modes.
For this purpose, are considering the possibility of defining the initial state $\sigma_0$
of the algorithm based on a scale~$S$ (second option below) rather than just its KS (first option below).
\def\optLead{\mathsf{opt}_\mathsf{lead}}
\def\optUnlead{\mathsf{opt}_\mathsf{unld}}
\begin{description}
\item[$(\optUnlead)$] 
$\sigma_0$ contains exactly the accidentals defined by the KS~of~$S$, 
\item[$(\optLead)$] 
$\sigma_0$ contains 
the accidentals of the KS of $S$ and 
the characteristic accidentals of~$S$.
\end{description}
\noindent
For the second option $(\optLead)$, in order to deal with non-diatonic scales like the blues modes, 
we extend the codomains of states from simple accidental into 
sets containing zero, one, or more accidentals.
\begin{example}
For the pentatonic $\Cp$ Major Blues scale, 
$\sigma_0(\Fp) = \sigma_0(\Bp) = \emptyset$, 
$\sigma_0(\Ep) = \{ \bemol, \becarre \}$, 
and  all other note names are mapped into the singleton set $\{ \becarre \}$.
\end{example}

The idea behind option $(\optLead)$ is that 
characteristic accidentals are important cues for guessing local scales, 
and they can be discounted.
Non-singleton sets of accidentals are only needed to build 
the initial state $\sigma_0$. Every update in a transition as above 
leads to a unique accidental.


\subsection{Naming inconsistencies and Chords}
\label{sec:chords} \label{sec:inconsistency}
\def\resSim{\mathsf{res}_\mathsf{sim}}
Let us consider the following restriction
that is not mandatory in music theory 
but turned out to be  important from a combinatoric point of view:
\begin{quote}
$(\resSim)$
two simultaneous notes 
(in the sense of $\inSimult$)
in the same pitch class must have the same name.
\end{quote}

Without the constraint $(\resSim)$, the presence of multiple chords within a single bar may result in
combinatorial explosion in some cases.

\begin{figure}
\begin{center}
\includegraphics[width=0.70\textwidth]{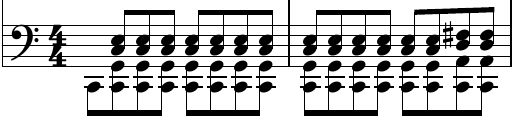}
\end{center}
\caption{Beethoven, Sonata 21 "Waldstein", measures 1-2, lh.} 
\label{fig:Waldstein}
\end{figure}

\begin{example}
At the beginning of the Waldstein Sonata, 
displayed in Figure~\ref{fig:Waldstein}, 
the treatment of simultaneous notes allows the algorithm not to treat all possible spellings 
for each occurence of doubled notes inside a chord. 
In this example assuming, $(\resSim)$, the $\Cp$'s in the repeated $\Cp$ major chord 
does not need to be spelled twice.
\end{example}

Although it is not a strict notational convention, 
counter examples to $(\resSim)$ are very rare.
We do not assume the other direction
(simultaneous notes with same name must be in the same pitch class),
as it can occur fairly frequently in tonal music, at least from the end of the nineteenth century, for instance in 
a dominant ninth chord with an appoggiatura on the ninth
(\eg $\Dp$ $\Fs$ $\Ap$ $\Cp$ $\Fn$ in $\Gp$ major), much appreciated by Ravel, among others.

The restriction $(\resSim)$ is ensured by adding to the states 
of Sections~\ref{sec:engraving}, \ref{sec:initialstate}
an additional partial mapping of  pitch classes into note names
$c \in \mathcal{C} = \{ \Ap,\ldots, \Gp \}^{\{ 0,\ldots, 11\}}$.
It is used to memorize the name associated to a pitch class
when processing a subsequence of simultaneous notes in $\bar{\nu}$.
In a state $\sigma$, when reading a simultaneous note of the subsequence 
with pitch class~$p$, 
we consider only transitions corresponding to a choice of note name~$n$
{compatible} with $c$ (in~$\sigma$), meaning that, 
if $c(p)$ is defined, then it is equal to $n$. 

\def\optInconsistency{\mathsf{opt}_\mathsf{inc}}
Moreover, we shall also consider the case of notes in the same bar
with the same pitch class and different names. 
We propose an option called $(\optInconsistency)$ when 
the number of such naming inconsistencies is added to the cost of transitions defined in
Sections~\ref{sec:engraving}, \ref{sec:initialstate}.


\section{Pitch Spelling Algorithm}\label{sec:algo}\label{sec:method}
We present a method 
addressing the problem presented in Section~\ref{sec:problem},
based on counting the accidentals 
that would be printed in a score 
following notational conventions recalled in Section~\ref{sec:shortest-path}
and using our extended Weber distance (Section \ref{sec:Weber}) between scales to refine its results.
The possible choices are exhaustively explored
through dynamic programming techniques.

Our approach, guided by common principles of musical notation and writing,
works in several steps: 
the first step (Section~\ref{sec:modal}) estimate one likely local scale for each bar, 
by considering the best spellings for each possible scale.
The second step (Section~\ref{sec:tonal}) uses the estimated local scales 
in order to refine the selection of KS and spellings. 
The local scales are essentially a byproduct of our algorithm, 
used in an intermediate state to estimate the best spellings.


\subsection{Modal Step}\label{sec:modal}\label{sec:step1}
%
From now on, assume we are given 
$p$ input notes to be spelled, distributed across 
$m$ bars, and 
$n$ candidate scales denoted $S_1,\ldots, S_{n}$.
We construct an 
$n \times m$ table~$G$, 
called the grid, where each entry~$G[i,j]$ assigns an estimated scale to bar~$j$, 
assuming the global starting scale is~$S_i$.
This involves two substeps.

\smallskip\noindent{\bf Substep 1}
is the construction of a $n \times m$ table~$T$, 
where~$T[i, j]$ is the cumulative cost of a best-path
computed with the algorithm of Section~\ref{sec:shortest-path} under the following conditions:
the initial state  $\sigma_0$ is defined by $(\optLead)$, 
without the option $(\optOctave)$, 
and, for each transition, a cost value defined by $(\optAccidCount)$ or $(\optAccidDist)$.
Thus,~$T[i, j]$ is roughly the number of accidentals printed $(\optAccidCount)$ 
or of accidental changes $(\optAccidDist)$, in bar $j$, 
under global scale~$S_i$.
%
When different best-cost paths arise, tie-breakers are used:
\begin{itemize}
\item the number of $\Cf$, $\Bs$, $\Ff$, $\Es$, 
\item the number of printed double accidentals not in the scale $S_i$, 
\item the number of accidentals not in the chromatic harmonic scale associated with $S_i$,
\item the number of printed accidentals with a sign different from the KS of~$S_i$:
        1 for $\bemol$, 2 for $\doublebemol$ when $\mbox{KS} > 0$, 
        and 1 for $\diese$, 2 for $\doublediese$ when $\mbox{KS} < 0$.
\end{itemize}

\smallskip\noindent{\bf Substep 2}
is the construction of the grid~$G$, 
where~$G[i, j]$ gives the estimated local scale for bar~$j$, assuming the starting scale is~$S_i$. 
\noindent
For $1 \leq i \leq n$ and $1 \leq j \leq m$, 
let $\mathsf{rk}_T(i, j)$  be the rank of $T[i, j]$ in the $j^\mathrm{th}$ column of $T$,
\ie the value $k$ such that global scale $S_i$ gives the $k^\mathrm{th}$ best cost in $T$ on bar $j$. 
Let $ \mathsf{rk}_\mathsf{dist}(S, S')$ be the rank of the distance $d(S, S')$,
as defined in Section~\ref{sec:Weber},
among $\langle d(S, S_1), \ldots, d(S, S_n)\rangle$.
Given $i_0  \in [1..n]$ the index of a hypothetical starting scale, 
let us consider the vector:
\[
\argmin_{i_0, i_1, \ldots, i_m \in [1..n]}
\left(
\sum_{j=1}^{m}  \mathsf{rk}_T(i_j, j)
+
\sum_{j=1}^{m}  \mathsf{rk}_\mathsf{dist}(S_{i_0},  S_{i_{j}})
+
\sum_{j=1}^{m-1}  \mathsf{rk}_\mathsf{dist}(S_{i_{j}},  S_{i_{j+1}})
\right)
\]
This yields, for every possible starting scale $S_{i_0}$ (candidate global scale), 
the sequence $\langle i_1, \ldots, i_m \rangle$ of local scale indices for each bar, 
minimising a combination of spelling cost 
(first term in the sum), 
distance to the global $S_{i_0}$ (second term), 
and modulation costs (third term).
The sequence is computed using a shortest-path algorithm 
similar to that of Section~\ref{sec:shortest-path}. 
Finally, we let $G[i_0, j] = S_{i_j}$.\\
A variant of this grid building process, which we call \textit{rank} and still use in some cases 
(see Section~\ref{sec:options}), 
is described in section 3.3.3 of~\cite{Bouquillard:hal-04458185}.


\subsection{Tonal Step}\label{sec:tonal}\label{sec:step2} 
The second and final step estimates a global KS ($\outKS$) 
and a note spelling in this KS  ($\outName$), using the local scales computed in grid~$G$. 

To this end, we construct a new $n \times m$ table $P$, similarly to $T$, 
but this time, with the options $(\optUnlead)$ and $(\optOctave)$ enabled. 
Additionally, the transition costs for computing $P[i, j]$ 
now account for the number of accidentals not in the estimated local scale $G[i, j]$. 
Intuitively, the spelling in $P[i, j]$ depends not only on the KS of $S_i$, 
but also on how well it fits the local scale $G[i, j]$.

More precisely, for each $i, j$, 
let $d_{i, j}$ be the number of accidentals in the transition that are not in $G[i, j]$ 
and let $a_{i, j}$ be the transition cost in $T[i, j]$ (Section~\ref{sec:modal}, Substep 1).
The transition cost is defined as a lexicographically ordered triple:
\begin{itemize}
\item[$i$.] $a_{i, j} + d_{i, j}$
\item[$ii$.] $d_{i, j}$
\item[$iii$.] the tie breakers of Section~\ref{sec:modal}.
\end{itemize}

The estimated global KS ($\outKS$)
is the Key Signature $K_i$ of $S_i$ 
where the row~$i$ of~$P$ is the one with the smallest cumulative cost. 
The chosen spelling ($\outName$) corresponds to the minimal-cost path in that same row~$i$.


\subsection{Deterministic Variant} 
\label{sec:PS13b}\label{sec:PS14}
We also propose a greedy variant of the algorithm presented 
in Sections~\ref{sec:step1} and~\ref{sec:step2}, 
which is more efficient but less exhaustive.
In this variant,  the choice of the spelling,  
with name~$n$ and accidental $a$ for the input note $\nu_i$ (see Section~\ref{sec:engraving}), 
is forced to the (unique) spelling 
in the chromatic harmonic scale of the current key~\cite{Nagel07chromatic}, 
instead of the 2 or 3 possible choices of Figure~\ref{fig:enharmonics}.
Hence, the transitions in the shortest path search are deterministic, 
\ie there is no need to search for best spelling in a bar because there is only one.
The rest of the algorithm works following the same steps described 
in Sections~\ref{sec:step1} and~\ref{sec:step2}.
The complexity of the table construction in this case is
$O(n \times p)$ where $p$ is the total number of notes in input and
$n$ is the number of keys considered.
This complexity is significantly better than the one of the above exhaustive algorithm.
In counterpart, some potentially correct spellings will be missed, 
as shown in the evaluation results presented below.

This approach is quite similar to the~{PS13} algorithm of~\cite{Meredith06ps13}, 
except that it uses the information on bars, 
which is assumed available in this paper but not in~\cite{Meredith06ps13}, 
in order to estimate global and local keys.
In~\cite{Meredith06ps13}, that estimation
is done (implicitely) by counting the number of occurrences  
of the (assumed) tonic note in a window whose
optimal size was evaluated manually.

\subsection{Rewriting Passing Notes} \label{sec:rewriting}
After note spelling has been chosen, we apply local corrections by rewriting passing notes, 
using a slight generalisation of 
the rules from D. Meredith’s PS13 pitch-spelling algorithm~\cite{Meredith06ps13}, step~2.
These rules follow classical voice-leading principles, 
we will discuss their relevance in a jazz context in Section~\ref{sec:evaluation}.

Each rule applies to a trigram of notes $\nu_0, \nu_1, \nu_2$,
separated by 1 or 2 semitones, in ascending, descending, 
or broderie patterns, 
where $\nu_0$ and~$\nu_1$, 
or $\nu_1$ and~$\nu_2$, 
share the same name.
In the application of a rule,  
the middle note $\nu_1$ is rewritten by changing its name. 
%
\newcommand{\rewrite}[5]{{#1} \overset{#2}{\mbox{---}} {#3} 
                              \overset{#4}{\mbox{---}} {#5} }
\begin{figure}
\[
\begin{array}{lrcl}
\mbox{broderie~down} & 
  \Cp\; \Cf\; \Cp & \to & \Cp\; \Bp\; \Cp\\
\mbox{broderie up} & 
  \Cp\; \Cs\; \Cp & \to & \Cp\; \Df\; \Cp\\
\mbox{descending}_{11} &
  \Cp\; \Cf\; \Ap & \to & \Cp\; \Bp\; \Ap\\
\mbox{descending}_{12} &
  \Cp\; \Cff\; \Af & \to & \Cp\; \Bf\; \Af\\
\mbox{descending}_{21} &
  \Cp\; \As\; \Ap & \to & \Cp\; \Bf\; \Ap\\
\mbox{descending}_{22} &
  \Cp\; \As\; \Af & \to & \Cp\; \Bf\; \Af\\
\mbox{ascending}_{11} &
  \Ap\; \As\; \Cp & \to & \Ap\; \Bf\; \Cp\\
\mbox{ascending}_{12} &
  \Af\; \As\; \Cp & \to & \Af\; \Bf\; \Cp\\
\mbox{ascending}_{21} &
  \Ap\; \Cf\; \Cp & \to & \Ap\; \Bp\; \Cp\\
\mbox{ascending}_{22} &
  \Ap\; \Cf\; \Cs & \to & \Ap\; \Bp\; \Cs\\
\end{array}
\]
\caption{Rewrite rules for passing notes (particular cases).}
\label{fig:rewrite}
\end{figure}

In Figure~\ref{fig:rewrite}, 
we present the rules for particular cases of notes. 
For instance, in the left-hand-side ${\Cp\, \Cf\, \Cp}$ 
of the first rule {broderie~down}, 
$\nu_0$, $\nu_1$, and $\nu_2$ all have the same note name~$\Cp$, 
the difference, in semitons, between $\nu_0$ and $\nu_1$ is $-1$ 
and the difference between $\nu_1$ and $\nu_2$ is $+1$. 
This rule rewrites the middle $\Cf$ ($\nu_1$) into~$\Bp$.

The rewrite rules are applied from left to right to the sequence spelled notes. 
At each rewrite step, at most one rule can be applied.


\section{Evaluation}\label{sec:eval}\label{sec:evaluation}

\subsection{Implementation}\label{sec:implem}\label{sec:code}
The algorithm of Section~\ref{sec:algo}
was implemented
in \textsf{C++}20 (17k loc).
This language was chosen for efficiency and integration into larger systems, 
in particular those designed for transcription, 
where quantised timings (especially bar boundaries) are computed before Pitch Spelling.
A Python binding, based on \textsf{\footnotesize pybind11}, 
offers calls (in Python) to the \textsf{C++} methods,
and was used for evaluation. 

For the evaluation, we used the Music21 toolkit~\cite{Cuthbert10music21},
to parse the ground-truth MusicXML 
score files in the evaluation datasets,  
extract the required note information (Section~\ref{sec:input}), 
and compare the estimated spellings and KS with those in the original scores. 
%
Some evaluation feedback is provided as 
tables (one row per opus) as well as output XML scores annotated with color-coded spelling differences, original spellings written under the staff when errors occur, and the estimated local scales and global KS (in grey)%
\footnote{The complete outputs of our evaluations on the 3 Jazz datasets are available at 
\url{https://github.com/florento/PSjazzEval}, and one can also find outputs obtained on classical datasets at \url{https://github.com/florento/PSEval}.}.

\subsection{Datasets}\label{sec:data}\label{sec:datasets}

We conducted an evaluation of our algorithm on 7 datasets,
based on the spelling in the original reference scores cited below (without annotations).

\subsubsection{Real Book.}
This dataset comprises 200 lead sheets of jazz standards  from the Real Book~\cite{RealBook}, 
in MusicXML format.
Some were digitised by us, others were sourced from MuseScore, 
and have been manually curated by us 
to conform to the reference edition~\cite{RealBook}, in both spellings and chord symbols.
Lead sheets are one page long on average, for a total of 
6000 bars and 21000 notes in the whole dataset. 

\subsubsection{Charlie Parker Omnibook.}
We consider a digitised version of~\cite{Deguernel16omax}, in musicXML format, consisting of 50 transcriptions
of complex tenor sax soli from the Charlie Parker Omnibook~\cite{Omnibook}%
\footnote{We  corrected a few spelling discrepancies between~\cite{Deguernel16omax} and the original~\cite{Omnibook}.}.
All scores in the dataset are transposed for C instruments. 
Some specificities of the scores in this dataset, regarding in particular note spelling, 
are discussed below in section~\ref{sec:options}.
The dataset contains a total of 3640 bars and 22700 notes.  

\subsubsection{FiloBass.}
We consider the dataset 
FiloBass~\cite{Riley23filobass} made of 48 MusicXML verified transcriptions of basslines of jazz standards,
whose backing tracks are digitised from the Aebersold series~\cite{AebersoldMethod},
obtained from the same backing tracks as used in the FiloSax dataset~\cite{Foster21FiloSax}, 
for a total of 12500 bars and 53000 notes. 
FiloBass was preferred to FiloSax as its XML files are public and it allowed to test on an different use case 
(basslines instead of saxophone soli, already present in Omnibook).

\subsubsection{The Session.}
We use a subset of traditional tunes extracted from the online community database The Session~\cite{TheSession}. The files, originally encoded in ABC format, were converted to MusicXML for our evaluation. This subset we spelled comprises 62 monophonic melodies. Such data interests us due to the modal nature of this folk repertoire. The subset contains a total of 7827 notes.

\subsubsection{ASAP.}
We have also used 5 separate corpora from the 222 pieces of the ASAP piano dataset~\cite{foscarin2020asap}
also in MusicXML format. 
All Bach preludes and fugues from the Well Tempered Clavier present in ASAP were used, 
except Preludes BWV 856 and 873 for technical reasons. 
All sonata movements by Mozart and Beethoven included in ASAP were also tested, 
as well as the K 475 Fantaisie by Mozart. 
Each of the 13 Chopin Etudes contained in ASAP, from both opus 10 and 25, 
was used, as well as the 8 Rachmaninov preludes present, from both opus 23 and 32. 
The cumulated total of notes spelled by our tested algorithms for this evaluation 
reaches a value of 216 464.

\subsubsection{DCML Schumann Kinderszenen.}
To further explore the Romantic piano repertoire, we also use of the Schumann \textit{Kinderszenen} Op.~15 corpus provided by the EPFL Digital and Cognitive Musicology Laboratory (DCML)~\cite{DCMLSchumann}. This dataset consists of the 13 short pieces comprising the opus, in MusicXML format. We relied strictly on the Pitch Spelling of the reference scores and ignored the Roman numeral annotations included in the DCML corpus. This dataset accounts for a total of 4,810 notes.

\subsubsection{La\-mar\-que-Goudard.}
We finally consider a monophonic (complex) dataset 
originated from the La\-mar\-que-Goudard rhythm textbook~\cite{Lamarque} 
\textit{D'un Rythme à l'Autre}, 
containing 250 excerpts, as MusicXML files, from pieces of extremely various styles, 
from Bach and Scarlatti to Wolf, Duparc, Debussy, Ibert...

\newcommand{\mini}[1]{{\small #1}}
\newcommand{\micro}[1]{{\tiny #1}}

\subsection{Evaluation Options and Ablation Tests}\label{sec:options}
We evaluated our algorithm using the options defined in Section~\ref{sec:algo} 
and compared its performances on the different corpora with those of several baselines: 
the default procedure for Pitch Spelling of the application MuseScore for score edition~\cite{musescore}, 
the data-driven model PKSpell from \cite{Foscarin21PKspell} 
and the famous Krumhansl-Schmuckler (K-S) key-finding model \cite{Krumhansl2001cognitive}. 

\paragraph{Jazz and Folk Datasets.}
Pitch Spelling results for the potentially modal repertoire such as Jazz and Folk are reported in Table~\ref{tab:JazzResults}. 
We vary the number of candidate scales for spelling 
(called $S_1,\ldots, S_n$ in Section~\ref{sec:algo}):
30 refers to major and harmonic minor modes for all KS, 
whereas 165 covers all the above, plus 6 other diatonic modes, melodic minor mode, 
and minor and major blues (see Section~\ref{sec:output}).
We evaluate with and without the passing note rewrite rules of Section~\ref{sec:rewriting} (post-processing).

All Jazz datasets 
include one or several Chord Symbols (CS) per bar in standard jazz notation~\cite{Levine95jazzTheory}.
We offer 2 options regarding these CS during spelling.
The first one, denoted by "no" 
in Table~\ref{tab:JazzResults},
 is simply to ignore them. 
The second option 
consists in extracting CS notes (via Music21~\cite{Cuthbert10music21}) and constraining their names during spelling.
In this "force" mode, at CS positions, only the transitions that match the CS note names are allowed in the shortest-path search (Section~\ref{sec:shortest-path}).
%
This last option makes sense in contexts such as automatic transcription of jazz soli, 
when the lead sheet is a standard whose CS are known in advance.

\paragraph{Classical Datasets.}
The algorithm presented in Section \ref{sec:algo} has also been tested on classical repertoire, 
both for Pitch Spelling and key estimation, 
against the several baselines cited above, with results presented in Table~\ref{tab:ClassicalResults}. 
On these corpora, our algorithm worked with 30 scales, all tonal, 
and used the \textit{rank} variant formula for its grid construction (see Section~\ref{sec:step1}). 
Performances of the deterministic variant of PSE (see Section~\ref{sec:PS14}), 
denoted by \mini{yes} in the "Deterministic" line, 
have also been assessed and reported in Table~\ref{tab:ClassicalResults} for each of the classical datasets. 
Since no equivalent of this variant exists for the baselines tested, 
the "Deterministic"  line in the MuseScore, PKSpell and Krumhansl-Schmuckler (K-S) columns is left blank.

\smallskip
In Tables \ref{tab:JazzResults} and \ref{tab:ClassicalResults}, 
the option combination (see Section~\ref{sec:engraving}) leading to the best results according to our recent experiments 
necessitated turning on $(\optUnlead)$ as well as $(\optAccidCount)$ for the path cost updates 
and turning off $(\optOctave)$ and $(\optInconsistency)$.

\paragraph{Editorial choices.}
In the Omnibook dataset~\cite{Riley24omnibook}, 
every opus uses a KS of 0, although the true global tonality is often not C major. 
We allow the option to force a global KS for $\outKS$ at the step in Section~\ref{sec:tonal}, but we did not use this option in our evaluations.
Moreover, this dataset does not contain
any  $\Bs$, $\Cf$, $\Es$, or $\Ff$, 
nor any double accidental~$\doublebemol$ or~$\doublediese$.
These can only be considered as editorial choices.
%
Therefore, we offer the option to disable such spellings in our algorithm by removing the corresponding transitions (as in Figure~\ref{fig:enharmonics}) in the shortest-path search (Section~\ref{sec:shortest-path}).
This editorial constraint is only applied to Omnibook as an option in our evaluations, 
as shown in the third result line dedicated to that dataset in Table~\ref{tab:JazzResults}.
It is not applied to the other datasets, which do contain such spellings.\\


\subsection{Results and Discussion}
Evaluation results for different options of our algorithm and various baselines 
on the 7 datasets described in Section~\ref{sec:datasets}
are displayed in Tables~\ref{tab:JazzResults} and \ref{tab:ClassicalResults}.

\begin{table*}
\caption{Accuracy of Pitch Spelling on Jazz and Folk datasets.}
\begin{center}
 \begin{tabular}{|c|c|c|c|c|c|c|c|c|c|c|c|c|c|c|c|c|c|c|}
 \hline
        & \mini{MuseScore} & \mini{PKspell} &
            \mini{PSE} & \mini{PSE} & \mini{PSE} & \mini{PSE} & \mini{PSE} & \mini{PSE} & \mini{PSE} & \mini{PSE}  \\
 \hline
 \hline
\mini{\bf scales} & & & \mini{30}   &  \mini{30}  &  \mini{30}  &  \mini{30} &  \mini{165} & \mini{165} & \mini{165} & \mini{165} \\
 \hline
\mini{\bf chords} & & & \micro{no} & \micro{no} & \micro{force} &  \micro{force} & \micro{no} & \micro{no} &  \micro{force} &  \micro{force} \\
 \hline
\mini{\bf rewrite} & &  & \micro{no} & \micro{yes} & \micro{no} & \micro{yes} & \micro{no} & \micro{yes} & \micro{no} & \micro{yes} \\
 \hline
 \hline
 \mini{\bf The Session} 
   & \mini{93.56}  & \mini{\bf 99.99} & \mini{99.86} & \mini{99.86} &*  &* & \mini{99.88} & \mini{99.88} & *&* \\
\hline
\mini{\bf Real Book} 
   & \mini{64.38} & \mini{96.71}  & \mini{97.35} & \mini{97.35}  & \mini{97.99} & \mini{\bf 98.01} & \mini{96.24} & \mini{96.77} & \mini{97.89}  & \mini{97.95} \\
\hline
\mini{\bf Omnibook}
  & \mini{85.66} & \mini{96.01}  & \mini{94.95} & \mini{94.88} & \mini{96.86} & \mini{ 96.78} & \mini{94.97} & \mini{94.88} & \mini{\bf 96.92} & \mini{96.85}  \\
\tiny{no $\Bs$, $\Cf$, $\Es$, $\Ff$, $\doublebemol$, $\doublediese$} 
  &                     & & \mini{97.23} & \mini{97.06}  &  \mini{97.91} & \mini{97.70} & \mini{96.78} & \mini{96.63} & \mini{\bf 98.04} & \mini{97.82} \\
\hline
\mini{\bf FiloBass} 
   & \mini{75.66} &  \mini{94.73} & \mini{94.68} & \mini{94.53} & \mini{94.84} & \mini{94.68} & \mini{94.87} & \mini{94.71} & \mini{\bf 95.23} & \mini{95.06} \\
\hline
\end{tabular}
\end{center}
\label{tab:JazzResults}
\footnotesize{* no chord symbol present in The Session subset, force mode thus not applicable}

\end{table*}

\begin{table*}[t]
\caption{Accuracy of Pitch Spelling and key estimation on classical datasets.}
\begin{center}
\begin{tabular}{|l|c|c|c|c|c||c|c|c|c|}
\hline
& note number & \multicolumn{4}{c|}{\mini{\bf Pitch Spelling (PS)}} & \multicolumn{4}{c|}{\mini{\bf Key Estimation (KE)}} \\
\hline
\mini{\bf Algorithm} & & \mini{MuseScore} & \mini{PKSpell} & \mini{PSE} & \mini{PSE} & \mini{K-S} & \mini{PKSpell} & \mini{PSE} & \mini{PSE} \\
\hline
\mini{\bf Deterministic} & & & & \mini{yes} & \mini{no} & & & \mini{yes} & \mini{no} \\
\hline
\hline
\begin{tabular}{l}
\mini{\bf Bach WTC}\\
\mini{\bf (ASAP)}
\end{tabular}
 & \mini{55,530} & \mini{94.13} & \mini{96.50} & \mini{98.27} & \mini{\bf 99.50} & \mini{87.27} & \mini{91.52} & \mini{98.29} & \mini{\bf 99.09} \\
\hline
\begin{tabular}{l}
\mini{\bf Mozart}\\ 
\mini{\bf(Fant. + Son.)}
\end{tabular}
 & \mini{13,830} & \mini{90.40} & \mini{\bf 99.20} & \mini{95.97} & \mini{97.65} & \mini{60.00} & \mini{\bf 80.00} & \mini{\bf 80.00} & \mini{\bf 80.00} \\
\hline
\begin{tabular}{l}
\mini{\bf Beethoven}\\
\mini{\bf (33 Son. mvt.)}
\end{tabular}
 & \mini{87,292} & \mini{91.91} &  \mini{\bf 97.81} & \mini{95.65} & \mini{97.64} & \mini{66.15} & \mini{90.48} & \mini{\bf 95.71} & \mini{92.32} \\
\hline
\begin{tabular}{l}
\mini{\bf Chopin}\\
\mini{\bf (13 \'Etudes)}
\end{tabular}
 & \mini{25,103} & \mini{91.19} & \mini{95.27} & \mini{96.03} & \mini{\bf 96.71} & \mini{84.62} & \mini{\bf 100} & \mini{96.15} & \mini{96.15} \\
\hline
\begin{tabular}{l}
\mini{\bf Schumann}\\
\mini{\bf (DCML)}
\end{tabular}
 & \mini{4,810} & \mini{86.90} & \mini{96.80} & \mini{96.82} & \mini{\bf 97.48} & \mini{61.54} & \mini{\bf 84.60} & \mini{79.16} & \mini{79.16} \\
\hline
\begin{tabular}{l}
\mini{\bf Rachmaninov}\\
\mini{\bf (4 Prel.)}
\end{tabular}
 & \mini{7,022} & \mini{84.30} & \mini{\bf 99.19} & \mini{97.49} & \mini{98.76} & \mini{\bf 100} & \mini{\bf 100} & \mini{\bf 100} & \mini{\bf 100} \\
\hline
\begin{tabular}{l}
\mini{\bf Lamarque}\\
\mini{\bf Goudard}
\end{tabular}
 & \mini{27,687} & \mini{96.49} & \mini{97.85} & \mini{98.23} & \mini{\bf 98.46} & \mini{50.60} & \mini{66.80} & \mini{74.30} & \mini{\bf 76.90} \\
\hline
\end{tabular}
\end{center}
\label{tab:ClassicalResults}
\end{table*}
We report a measure of accuracy, \ie the percentage of estimated spellings
that conform to those of the XML scores from the datasets.
We take into consideration grace notes, and notes in chords 
(there are few chords in the jazz evaluation datasets).
Tied notes are counted only once, \ie we ignore the spelling 
of a note tied to a previous one.
Of course, when using the option to "force" chord symbols, 
we do not count the spelling of the notes in these chord symbols. 
In Table \ref{tab:JazzResults}, we do not give the results of KS estimation, 
which are not relevant in this context.
Let us recall that KS and local scale estimations are by-products of the algorithm, 
mostly necessary for computing spellings.

We ran MuseScore \cite{musescore} XML to MIDI and then MIDI to XML conversion command to evaluate its built-in Pitch Spelling quality, as well as PKspell~\cite{Foscarin21PKspell} 
on evaluation datasets (first and second column of Tables~\ref{tab:JazzResults}) and \ref{tab:ClassicalResults}).
Note that the options on the number of tonalities, 
rewriting as well as exclusion of $\Bs$, $\Cf$, $\Es$, $\Ff$ and double accidentals
are not relevant in the case of MuseScore nor PKspell.
The chord symbols were discarded (not spelled) for the evaluation with MuseScore and PKspell.

The algorithm of Section~\ref{sec:method} is called PSE in Table~\ref{tab:JazzResults}.
The input notes are supplied using the values 
$(\inPitch)$, $(\inSimult)$ and $(\inBar)$ presented in Section~\ref{sec:input}.
For PKspell, the input is made of the pitch class and duration (in Music21 \textit{quarterlength}~\cite{Cuthbert10music21}) 
of all notes in the datasets in the order in which they appear. 

\paragraph{Discussion Jazz repertoire}
We observe different trends in the results obtained with PSE on the Real Book and Omnibook (Table~\ref{tab:JazzResults}).
The shift from 30 to 165 scales 
generally improves spelling scores for the Omnibook, 
while it worsens them for the Real Book.
As for the rewriting of passage notes, overall,  
the situation is the opposite.
This may be due to the different nature of the datasets.
On the one hand, Real Book standards are made of rather simple melodies, fitting within the 30 tonal scales. 
The behaviour of PSE with Real Book standards \wrt rewriting is somewhat similar 
to what is observed on the classical Datasets in Table~\ref{tab:ClassicalResults} 
with classical monophonic and piano datasets. 
On the other hand, the Omnibook contains complex improvisations, 
with frequent key changes and implied harmonies, 
often drawing on a wider range of scales (beyond the 30-note scale, into the 165-note scale), even for brief forays,
and featuring a high degree of chromaticism that sometimes renders traditional transposition rules inadequate.

\noindent 
As one might expect, activating the option to exclude $\Bs$, $\Cf$, $\Es$, $\Ff$ and double accidentals
improves drastically the spelling results on the Omnibook dataset.

The evaluations gave poorer results on the FiloBass corpus than on the others.
As a potential explanation, we can point out an inherent difficulty of tackling the Pitch Spelling problem in a jazz context, 
which is the great variability of spelling logics in the available transcription datasets, sometimes even within a single piece. 
For instance, in the beginning of the 14th piece of the FiloBass corpus, 
\begin{figure}
\begin{center}
\begin{tabular}{ccccccc}
\raisebox{10pt}{\includegraphics[width=0.13\textwidth]{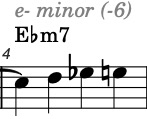}} & 
\raisebox{21pt}{\ldots} & 
\includegraphics[width=0.13\textwidth]{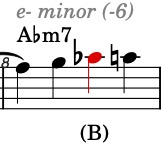} & $\qquad\qquad$ & 
\includegraphics[width=0.13\textwidth]{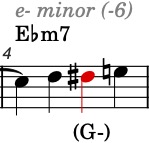} &  
\raisebox{21pt}{\ldots} & 
\raisebox{11pt}{\includegraphics[width=0.115\textwidth]{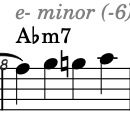}} \\
\end{tabular}
\end{center}
\caption{Different spellings of bars~4 and~8 in opus~14 of FiloBass; bass clef, {KS} = $-1$, 
        original spellings shown under the staff, estimated local scales written above it in grey.}
\label{fig:Four}
\end{figure}
the Hard-Bop standard \emph{Four} of Eddie Vinson and Miles Davis, 
the 4th and 8th bars can be fruitfully compared - see Figure~\ref{fig:Four}. 
Although they constitute a simple transposition by a Perfect Fourth of one another, 
the choices of spelling diverge for the third note of each bar.
In the 4th bar, the surrounding harmonic context of $\Ef$\,\textsf{m7} leads to a~$\Gf$, 
which is correctly predicted by our algorithm 
when using the cost class presented in Section~\ref{sec:tonal}.
This cost combines the number of accidentals to the distance to the estimated local scale (left in Figure~\ref{fig:Four}). 
In the 8th bar however,  the reference chooses the other possibility~$\Bn$ instead of~$\Cf$.
This alternative spelling is predicted by our algorithm if we change the cost class in the Tonal Step of Section~\ref{sec:tonal}
to a lexicographically ordered tuple: first  the number of printed accidentals, then the distance to the current local scale. In this case, however, bar 4 is no longer spelled correctly (right in Figure~\ref{fig:Four}). 

\paragraph{Discussion classical repertoire.}
In Table~\ref{tab:ClassicalResults},
looking at global and local tonality estimation, and although it is beaten by PKSpell on Schumann and Chopin, PSE achieves good results, 
especially when comparing Key Signatures together. 
It consistently and by a considerable margin outperforms the Krumhansl-Schmuckler (K-S) 
key-finding algorithm~\cite{Krumhansl2001cognitive}, 
which calculates, for each major and minor key, 
a correlation coefficient between the values of the tested key’s profile and the total durations of the corresponding pitch class in the given piece of music, and then selects the best key based on the calculated correlation coefficients. 
It is interesting to note that our algorithms need only know the boundaries of the measures and not the note durations, 
whereas the KS algorithm uses note durations and does not take measures into account. 

We have observed that the version of PSE using 30 scales tends to favour minor keys over their major relatives.
Minor keys do indeed contain a greater number of notes, 
especially if the ($\optLead$) option is selected: 
no penalty will affect the notation, whether in harmonic or natural minor modes, 
or in the ascending melodic mode. 
In this configuration, the only error of global tone estimation on our ASAP Well-Tempered Clavier dataset is directly due to this tendency: 
the presence of natural $\Bp$'s in the BWV~870 prelude ($\Cp$ major of book~2), 
in a piece where flat $\Bp$s are also numerous due to modulations to $\Fp$ major, 
$\Dp$ minor \etc., did not prevent our algorithm 
from estimating the piece as written in $\Dp$ minor, 
because these natural $\Bp$'s were interpreted as part 
of the ascending melodic minor mode of $\Dp$, 
instead of indicators of a $\Cp$ major context. 
This issue is addressed in the current extended version of PSE: 
when working with 165 scales, one individual scale represents each minor mode, 
thus making the asymmetry between a more tolerant and plural minor mode versus a more restrictive major mode disappear.

\paragraph{General discussion.}
From both Tables~\ref{tab:JazzResults} and~\ref{tab:ClassicalResults}, 
It seems clear that the PS algorithm built into MuseScore performs the worst in terms of Pitch Spelling
among the programs tested on all datasets,
even though it was evaluated under more favorable conditions than PSE and PKSpell~\cite{Foscarin21PKspell}.
In fact, we allowed it to take the ground truth Key Signature into account 
(the KS is copied when converting scores to MIDI format), 
since the program was not designed to guess a piece’s KS; 
so MuseScore could use this additional information to refine the spelling correction, 
which neither PSE nor PKSpell can do, since they aim to predict both the KS and the spelling of notes. 

The comparison of the results of PKspell~\cite{Foscarin21PKspell}
and PSE is less significant and should in any case be taken with more caution.
As explained before, input is not the same for both tools, 
in particular regarding the timings.
Moreover, 
the model of PKspell was trained  with the dataset ASAP~\cite{foscarin2020asap} of classical piano music.
We 
do not know whether re-training it with jazz data would improve 
the results on the datasets of Table~\ref{tab:JazzResults}. It is worth noting that the version of PSE without the jazz modes obtains on average
better results than PKspell on the datasets ASAP, DCML and Lamarque-Goudard shown in Table \ref{tab:ClassicalResults}
in an order of magnitude similar to the improvement achieved on the Real Book in Table~\ref{tab:JazzResults}, although serious contrasts can be observed across composers, with Mozart for instance being significantly better spelled ($>1.5\ \%$ difference) by PKSpell than PSE.

%



\section{Conclusion}\label{sec:conclusion}
We have presented a Pitch Spelling algorithm
that takes as input time-ordered MIDI pitches with bar boundaries
and estimates an global Key Signature, a local scale per bar,
as well as note names based on these estimates.
Our algorithm relies both on engraving rules, 
following a principle of parsimony used in Common Western Music Notation 
to print accidentals in a score, 
and on the type of reasoning used by musicians when they deduce local keys from accidentals. 
The algorithm’s design choices are guided by a concern for musical relevance.

It was evaluated on various types of datasets:
jazz lead sheets, 
transcriptions of complex improvised tenor saxophone soli and jazz bass lines, 
folk tunes, 
as well as sheet music for classical piano and other monophonic instruments.
The results are generally good, but vary from one dataset to another, 
reflecting the diversity of the repertoires.
In the case of jazz transcriptions, certain inconsistencies 
in notation conventions across different sources 
also complicate the evaluation process.
This variability in results suggests that there is no one-size-fits-all approach 
and justifies the use of the various options presented in this article
for different use cases.

Our algorithm ignores the duration of entered notes, 
except when it comes to bar boundaries.
This is a design choice intended to ensure maximum generality.
Taking this information into account could be the subject of further experimental studies.
However, this would require establishing a relationship between durations 
and the cost values in Section~\ref{sec:method}, 
a task that does not seem easy to calibrate in a non-arbitrary manner. 
For example, we are not convinced that the assumption that longer notes 
are more likely to be genuine chordal factors 
and should therefore carry more weight in the chord notation —
which is generally true in the classical repertoire —
still holds true in the context of improvised jazz music.

The experimental results above compare the advantages 
of an algorithmic approach like ours 
and a data-driven approach such as ~\cite{Foscarin21PKspell}
for Pitch Spelling, particularly in the field of jazz. 
One of the main advantages 
of trained statistical models in this context 
is that they relieve the designer of the need to manually adjust cost functions and parameters.
The advantages of an algorithm such as the one presented in this article 
are the advanced control it offers users, 
with options such as those presented above, 
e.g., for editorial choices that may be related to an instrument or style, 
as well as the explainability of the results, 
since each spelling decision can be linked to explicit and interpretable rules.

\begin{credits}
\subsubsection{\ackname} 
This work has been partially funded by a grant from the Inria Exploratory Action Codex.
The authors would like to thank John Xavier Riley for providing them with the dataset FiloBass, 
Ken Deguernel for his valuable advice about jazz transcription, 
and Jean-Paul Despax for the enlightening conversations he shared with us.

\subsubsection{\discintname}
The authors have no competing interests to declare that are relevant to the content of this article. 
\end{credits}
%
%
\bibliographystyle{splncs04}
\bibliography{references}

\begin{thebibliography}{10}
\providecommand{\url}[1]{\texttt{#1}}
\providecommand{\urlprefix}{URL }
\providecommand{\doi}[1]{https://doi.org/#1}

\bibitem{Omnibook}
Aebersold, J., Slone, K.: {C}harlie {P}arker {O}mnibook. Hal Leonard (1978)

\bibitem{AebersoldMethod}
Aebersold, J.: Jamey {A}ebersold {J}azz play-{A}-{L}ong {S}eries (2000),
  \url{http://jazzbooks.com/jazz/JBIO}

\bibitem{Bouquillard:hal-04458185}
Bouquillard, A., Jacquemard, F.: {Engraving Oriented Joint Estimation of Pitch
  Spelling and Local and Global Keys}. In: International Conference on
  Technologies for Music Notation and Representation (TENOR). Zurich University
  of the Arts (2024)

\bibitem{Cambouropoulos03pitch}
Cambouropoulos, E.: Pitch spelling: A computational model. Music Perception
  \textbf{20}(4),  411--429 (2003)

\bibitem{Chew05spiral}
Chew, E., Chen, Y.C.: Real-time pitch spelling using the spiral array. Computer
  Music Journal  \textbf{29}(2),  61--76 (2005)

\bibitem{Cuthbert10music21}
Cuthbert, M.S., Ariza, C.: music21: A toolkit for computer-aided musicology and
  symbolic music data. In: 11th International Society for Music Information
  Retrieval Conference (ISMIR) (2010)

\bibitem{Deguernel16omax}
D{\'e}guernel, K., Vincent, E., Assayag, G.: Using multidimensional sequences
  for improvisation in the {OM}ax paradigm. In: 13th Sound and Music Computing
  Conference (SMC) (2016)

\bibitem{TheSession}
Effort, C.: The session. \url{https://thesession.org} (2001), accessed: 2026

\bibitem{Feisthauer20smc}
Feisthauer, L., Bigo, L., Giraud, M., Lev{\'e}, F.: Estimating keys and
  modulations in musical pieces. In: SMC (2020)

\bibitem{Foscarin21PKspell}
Foscarin, F., Audebert, N., Fournier-S'Niehotta, R.: {PKS}pell: {D}ata-driven
  pitch spelling and key signature estimation. In: ISMIR (2021)

\bibitem{foscarin2020asap}
Foscarin, F., Mcleod, A., Rigaux, P., Jacquemard, F., Sakai, M.: {ASAP}: {A}
  dataset of aligned scores and performances for piano transcription. In: ISMIR
  (2020)

\bibitem{Foster21FiloSax}
Foster, D., Dixon, S.: A dataset of annotated jazz saxophone recordings. In:
  ISMIR (2021)

\bibitem{Gould11Notation}
Gould, E.: {B}ehind {B}ars: {T}he definitive guide to music notation. Faber
  Music (2011)

\bibitem{DCMLSchumann}
Hentschel, J., Rammos, Y., Moss, F., Neuwirth, M., Rohrmeier, M.: An annotated
  corpus of tonal piano music from the long 19th century. Empirical Musicology
  Review  \textbf{18},  84--95 (01 2024). \doi{10.18061/emr.v18i1.8903}

\bibitem{Honingh09compactness}
Honingh, A.K.: Compactness in the {E}uler-lattice: A parsimonious pitch
  spelling model. Musicae Scientiae  \textbf{13}(1),  117--138 (2009)

\bibitem{Huang08advanceddynamic}
Huang, L.: Advanced dynamic programming in semiring and hypergraph frameworks.
  In: COLING (2008)

\bibitem{Krumhansl2001cognitive}
Krumhansl, C.L.: Cognitive foundations of musical pitch, vol.~17. Oxford
  University Press (2001)

\bibitem{Lamarque}
Lamarque, E., Goudard, M.J.: D'un Rythme {\`a} l'autre, vol.~1-4. Lemoine
  (1997)

\bibitem{RealBook}
Leonard, H. (ed.): The Real Book. Hal Leonard, 6th edn. (2007)

\bibitem{Levine95jazzTheory}
Levine, M.: The Jazz Theory Book. Sher Music Co (1995)

\bibitem{Meredith06ps13}
Meredith, D.: The {PS13} pitch spelling algorithm. Journal of New Music
  Research  \textbf{35}(2),  121--159 (2006)

\bibitem{musescore}
{MuseScore BVBA}: {MuseScore}: Free music composition and notation software.
  \url{https://musescore.org} (2023), version 4.x

\bibitem{Nagel07chromatic}
Nagel, J.: The chromatic modal scale: Proper spelling for tonal voice-leading.
  JOMAR Press  (2007)

\bibitem{Riley23filobass}
Riley, X., Dixon, S.: Filobass: A dataset and corpus based study of jazz
  basslines. In: 24th Int. Society for Music Information Retrieval Conference
  (ISMIR) (2023)

\bibitem{Riley24omnibook}
Riley, X., Dixon, S.: Reconstructing the {C}harlie {P}arker {O}mnibook using an
  audio-to-score automatic transcription pipeline. In: SMC (2024)

\bibitem{Temperley04cognition}
Temperley, D.: The cognition of basic musical structures. MIT press (2004)

\bibitem{Teodoru07pitch}
Teodoru, G., Raphael, C.: Pitch spelling with conditionally independent voices.
  In: ISMIR (2007)

\bibitem{Weber}
Weber, G.: Versucht einer geordneten Theory der Tonsetzkunst. B. Schott's
  Sohnen (1818)

\bibitem{Wetherfield20minimum}
Wetherfield, B.: The minimum cut pitch spelling algorithm: Simplifications and
  developments. In: TENOR (2020)

\end{thebibliography}
%
%
%
%
%

\clearpage
\onecolumn
\appendix
\pagenumbering{alph}


%

\end{document}